\documentclass[5p,times,authoryear,twocolumn]{elsarticle} 

\usepackage{amsmath}
\usepackage{graphicx}
\usepackage{lipsum}
\usepackage{subcaption}        
\usepackage{lscape}           
\usepackage{placeins}           
\usepackage{tabularx}
\usepackage{float}
\usepackage{booktabs}
\usepackage{ragged2e}           
\usepackage{subdepth}
\usepackage{hyperref}
\usepackage{xurl}
\usepackage{lineno}
\usepackage{setspace}

\newcolumntype{Y}{>{\Centering\arraybackslash}X}
\newcolumntype{P}[1]{>{\centering\arraybackslash}p{#1}}

\begin{document}
% \linenumbers
\begin{frontmatter}

    % --- TITLE ---
    
    \title{A Multispectral Framework for the Detection of Calcium Carbide-Induced Ripening and Shelf-Life Estimation in Climacteric Fruits}

    % --- AUTHORS ---
  
    \author[1]{Gurbhit Chaurakoti}
    \author[1]{Harshit Kumar}
    \author[1]{Hani Kumar}
    \author[2]{Anurag Singh\corref{cor1}}
    \author[3]{Ram Asrey}

    \cortext[cor1]{Corresponding author. Email: anuragsg@nitdelhi.ac.in}

    \affiliation[1]{organization={Department of Electrical Engineering, National Institute of Technology Delhi}, city={New Delhi}, postcode={110036}, country={India}}
    
    \affiliation[2]{organization={Department of Computer Science and Engineering, National Institute of Technology Delhi}, city={New Delhi}, postcode={110036}, country={India}}

    \affiliation[3]{organization={Division of Food Science and Post Harvest Technology, ICAR-Indian Agricultural Research Institute}, city={New Delhi}, postcode={110012}, country={India}}
    
    % % --- ABSTRACT ---
    \begin{abstract}
    Significant health risks are associated with the illegal, yet commonly practiced use of industrial-grade Calcium Carbide (CaC\textsubscript{2}) for ripening of climacteric fruits like mango and banana, which leaves behind trace residues of arsenic and phosphorus.
    To address this, the proposed study explores a novel, non-invasive multispectral framework for distinguishing safely ripened fruits (naturally ripened and ethephon-induced) from calcium carbide-ripened samples, while also estimating their ripening progression (in percentage) and remaining shelf life (in days).
    The spectral profiles of Mango (Mangifera indica) and Banana (Musa acuminata) at 18 discrete wavelengths in the visible-near infrared (NIR) range (410 nm – 940 nm) are studied using the AS7265x spectral triad sensor. 
    CaC\textsubscript{2} treated samples exhibit sharper spectral intensity drops in the visible region wavelengths, consistent with accelerated chlorophyll 
    degradation and carotenoid development. To characterize these physiological changes, the feature engineering strategy integrates inter-method spectral variance, intensity ratios at distinct wavelengths, and environmental parameters including temperature and humidity. Dimensionality reduction using Principal Component Analysis (PCA) retains >90\% of spectral variance within the first 5–7 components. The resulting feature set is used to train three independent eXtreme Gradient Boosting (XGBoost) based learning algorithms for ripening method classification along with quantitative estimation of remaining shelf life and ripening progression. A classification accuracy of 95\% along with carbide class recall of 0.67 is observed for mango samples, while the model achieves an accuracy of 81\% and carbide 
    class recall of 0.74 for banana. This instrumentation and data-driven approach demonstrates the effectiveness of the proposed non-invasive framework.
  \end{abstract}

    % --- KEYWORDS ---

\end{frontmatter}

% \section{Introduction}

%%%%%%%%%%%%%%%%%%%%%%%%%%%%%%%%%%%%%%%%%%%%%%%%%%%%%%%%%%%%%%
\section{Introduction}

Mango and banana are among the most consumed fruit crops globally \citep{heng2017}. Both of these fruits are climacteric, which means that they continue to mature even after being picked from the trees. Consequently, to minimize damage during transit and meet market demand, the application of artificial ripening agents post-harvest is often required. Calcium Carbide (CaC\textsubscript{2}) is widely used for this purpose due to its low cost and easy availability \citep{Vidhya2025}.
However, CaC\textsubscript{2} poses serious health risks due to the presence of trace impurities such as arsenic (As) and phosphorus (P), leading to a ban on its use as a ripening
agent for fruits in most countries across the world, including Malaysia (Food Act 1983), Sri Lanka (Food Act No 26, 1980), the Philippines (Food Safety Act, 2013) and India (Prevention of Food Adulteration Act, 1955 and the Food Safety and Standards Regulations, 2011) \citep{Vidhya2025}. Yet, an increase in the use of industrial grade CaC\textsubscript{2} for fruit ripening has been reported in recent times \citep{OKEKE2022153112}. To address this challenge, the proposed study explores a non-invasive framework using visible-near infrared spectroscopy to distinguish calcium carbide-ripened mango and banana samples, from naturally ripened ones, along with estimating the remaining shelf life of the fruit.

The natural ripening of climacteric fruits is a process driven by the plant hormone ethylene (C\textsubscript{2}H\textsubscript{4}). Ethylene starts a series of enzymatic reactions that lead to the transition of fruit from an unripe state to one ready for consumption. Key changes include the breakdown of chlorophyll pigments, which reveals underlying carotenoids such as xanthophylls and carotenes. Consequently,
noticeable changes in the spectral profile of fruits are also observed \citep{Chidangil2017}.
A safe method of postharvest ripening, also permitted by the Food Safety and Standards Authority of India (FSSAI) \citep{fssai2020}, is controlled exposure of fruits to ethylene gas in ripening chambers that maintain an optimal temperature (15°–25°C) and relative 
humidity (90–95\%), with ethylene concentration not exceeding 100 ppm. \citet{maduwanthi2021} reported that liquid ethephon (2-chloroethylphosphonic acid); also known as ethrel breaks down into 
ethylene when mixed with water. Ethephon sachets are often used as exogenous ripening agents for raw fruit samples, especially during transportation. However, CaC\textsubscript{2} is more commonly
used to induce post-harvest ripening in fruits. CaC\textsubscript{2} when in contact with moisture, produces acetylene (C\textsubscript{2}H\textsubscript{2}); which is an analogue of ethylene and mimics its effect on fruits. The acetylene gas produced from CaC\textsubscript{2} contains 
traces of phosphine and arsine up to 95 and 3 ppm, respectively \citep{siddiqui2010,maduwanthi2019}. While acetylene effectively induces ripening, it simultaneously leads to the accumulation of arsenic, phosphorus, and other toxic heavy metals, such as lead (Pb), cadmium (Cd), iron (Fe), and mercury (Hg), within the fruit tissues. Consumption of arsenic is linked to serious health problems \citep{hong2014} including higher chances of various cancers (skin, lung, liver, bladder), 
skin lesions, heart disease, nerve disorders, kidney damage, and issues in immune system. Arsenic poisoning leads to stomach pain, weakness, trouble swallowing, numbness, low blood pressure, and can be lethal. A study by \citet{stat} on mangoes found a cancer risk (CR) value of $1.4 \times 10^{-3}$ for arsenic, which exceeds the acceptable upper limit of $1.0 \times 10^{-4}$, indicating that the consumption of CaC\textsubscript{2} ripened mangoes poses carcinogenic risk. High quantities of phosphorus can upset mineral balance, resulting in bone loss, kidney problems, and artery calcification \citep{richard2022}. Moreover, the acetylene gas produced during the ripening process acts as a nervous system depressant. Inhalation of these byproducts can induce
headaches, dizziness, mood changes, memory loss, and seizures. Workers handling calcium carbide are at risk too, facing possible respiratory issues, skin and eye irritation, or burns from direct contact. Despite the legal implications and 
severe health risks, CaC\textsubscript{2} remains readily available and extensively used for artificial ripening in the Indian subcontinent, parts of Africa and South America \citep{rasdi2018}, which makes it crucial to explore and develop methods 
for detecting the use of CaC\textsubscript{2} in fruit ripening.

Methods for detecting the use of CaC\textsubscript{2} in fruit ripening have evolved
from very sensitive lab-based techniques to faster, non-invasive
technologies. Traditional methods, like HS-SPME-GC-MS \citep{gcms} and
electrochemical biosensors \citep{biosenseor}, are very specific and sensitive.
They can identify unique volatile markers, such as 3,5
dimethyl-1,2,4-trithiolane, or measure CaC2 residues at the
nanomolar level. However, their complexity, expense, and
invasive nature limits their use in large-scale monitoring.
As a result, the field is moving toward non-invasive solutions. Techniques based on detecting impurities, like the AuNP-based
colorimetric assay \citep{lakade2018}, have shown effectiveness by using arsenic
as a reliable marker. NIR spectroscopy based approaches
have proved to be effective \citep{Lakade2019NIRSM,Macadaeg2025NIRLED}, especially the use of intensity
ratios across wavelengths \citep{Ifmalinda2025Avocado}, that represent ripening progression
and biochemical markers of calcium carbide-ripening.\\

\subsection{\textbf{Background}}

The proposed study is grounded in established frameworks for data characterization and predictive modelling. It leverages some of the widely utilized statistical techniques for data analysis, augmentation and feature engineering. The hardware includes the ESP32 microcontroller, SHT40 temperature and humidity sensor and AS7265x spectral sensor. The ESP32 is a powerful System on Chip (SoC) for Internet of Things (IoT) and embedded applications, widely used for experimental studies. The SHT40 is a digital sensor that provides accurate measurements of temperature and relative humidity, while consuming minimal power and communicating via an Inter-Integrated Circuit (I²C) interface. The AS7265x spectral sensor module has three integrated sensors (AS72651, AS72652, and AS72653). They together offer 18 distinct spectral channels ranging from 410 nm (violet in visible spectrum) to 940 nm (NIR spectrum) for measuring spectral intensity at the corresponding wavelengths, making it suitable for spectral analysis.
 
In the proposed study, ratio of spectral intensities at distinct wavelengths, principal components, z-scores, temperature and humidity are identified as potential input features for the learning algorithm. Principal Component Analysis (PCA) is used for reducing dimensionality of spectral data by converting the original feature space with higher dimensions into a new set of orthogonal components that capture majority of spectral variance in fewer dimensions \citep{vidal2005,jolliffe2002}.Z-score indicates the number of standard deviations by which a data point differs from the mean \citep{yan2025}.
Tools such as Spearman’s rank correlation, Interquartile Range (IQR), Analysis of Variance (ANOVA), and SHapley Additive exPlanations (SHAP) are used in the study for identifying relevant patterns in the data and establishing a robust input feature set for training the learning algorithm. Spearman's rank correlation measures the strength and direction of monotonic association between two rank variables. Spearman's rank correlation coefficient ranges from $-1$ to $+1$, where $+1$ denotes a perfect positive monotonic association, $0$ indicates no monotonic association between the ranked variables, and $-1$ denotes a perfect negative monotonic association \citep{wu2022}. Since median is not affected by outliers, it provides a reliable measure of central tendency for skewed or non-normally distributed data. In such cases, IQR is used to describe the data dispersion. IQR quantifies the spread of the middle 50\% of observations from first quartile ($Q_1$) to third quartile ($Q_3$) \citep{aznargimeno2023} which is calculated as, 
\begin{equation}
IQR = Q_3 - Q_1
\end{equation}

Intersection over Union (IoU), also known as the Jaccard Index, is a statistical metric used to quantify the degree of overlap between two sets or distributions \citep{MOHAMMADI2025100624}.
In the context of assessing an input feature in the proposed study, let $A$ and $B$ represent the IQRs of a specific feature for two distinct ripening methods. The IoU is mathematically defined as the ratio of the magnitude of their 
intersection to the magnitude of their union:

\begin{equation}
\text{IoU}(A, B) = \frac{|A \cap B|}{|A \cup B|}
\end{equation}

Where:
\begin{itemize}
    \item[] $|A \cap B|$ represents the Intersection: The length of the interval where the feature values for both ripening methods overlap.
    \item[] $|A \cup B|$ represents the Union: The total span covered by both distributions, calculated as $|A| + |B| - |A \cap B|$.
\end{itemize}

This metric provides a normalized value between 0 and 1, where $\text{IoU} \to 0$ indicates high class separability, and $\text{IoU} \to 1$ indicates a high degree of overlap.\\\\
ANOVA is a statistical technique used to determine whether there are significant differences between the means of independent groups based on a single categorical factor. A one-way ANOVA evaluates the variance between these groups as well as within these groups to assess how the independent variable affects a continuous dependent variable. 

The procedure involves testing a null hypothesis ($H_0$), which is defined as:
\begin{equation}
    H_0: \mu_1 = \mu_2 = \dots = \mu_k
\end{equation}
where $\mu$ represents the population mean of each group. This hypothesis implies that the independent categorical factor has no significant effect on the continuous dependent variable \citep{Kwak2012}.
To evaluate these hypotheses, a \textit{p}-value is calculated, which is defined as the probability of obtaining test results at least as extreme as the observed results, assuming the null hypothesis is true. If the calculated \textit{p}-value is less than or equal to the significance level (e.g., $p \le 0.05$), the null hypothesis is rejected. This indicates that the mean values are significantly different and there is a significant effect of the independent categorical factor on the dependent variable.\\\\
SHAP is used to understand how the model makes decisions by measuring the impact of each input feature on the model’s output. SHAP assigns an importance value to each feature, which reflects how much that feature increases or decreases the predicted outcome as compared to the model’s average prediction or its baseline value. Before labeling the output, model generates continuous output values, called logits or probabilities. SHAP works on these logits instead of the output labels. By comparing the SHAP values, the influence of each feature on the model’s output can be determined to rank them accordingly based on their overall importance \citep{wu2025}.

In the proposed study, augmentation is performed using noise injection to increase the variability of the training set by creating altered versions of existing samples. Multiplicative Gaussian noise is applied such that each sample is modified as $x_{\text{aug}} = x \cdot \epsilon$. Here, $x$ represents the original value, $x_{\text{aug}}$ is the augmented value, and $\epsilon$ is a random value of noise taken from a normal (Gaussian) distribution \citep{beddiar2023}, where $\epsilon$$\sim$$\mathcal{N}(\mu, \sigma^2)$.

XGBoost is a gradient boosting algorithm that builds an ensemble of decision trees such that each new tree tries to correct the mistakes made by previous trees. It uses regularization to avoid overfitting. Grid Search Cross-Validation (GridSearchCV) is used for hyperparameter tuning to optimize the model. It tests each combination of given hyperparameters and finds the combination that performs best during cross-validation \citep{duan2024}.

\subsection{\textbf{Literature Survey}}

Digital image processing represents one of the most accessible and cost-effective techniques for detecting calcium carbide-induced ripening. Smartphone cameras are used with computer vision algorithms to analyse surface-level features \citep{smartphone}. \citet{sreeram2025} used computer vision and deep learning on 
RGB images captured by Pi camera using Raspberry Pi in real time to detect chemically ripened mangoes and bananas. Among chemical methods, colorimetric impurity-based tests provide a portable, 
real-time and low-cost approach, making them suitable for large-scale field inspection and even consumer-level use. For instance, gold nanoparticle (AuNP)-based assays \citep{lakade2018} offer a reliable visual metric, transitioning in color from red to purple in the presence of chemical markers indicative of CaC\textsubscript{2} ripening. Electrochemical biosensors and e-noses can be used in handheld devices for field applications, where volatile organic compound (VOC) patterns can be used to train machine learning models. \citet{ghatak2021} developed a low-cost, portable gas sensing system for identifying artificial ripening in mangoes. Hyperspectral imaging is the most accurate non-invasive method which captures spectral data for each pixel in an image, 
providing a large volume of data to analyse and find relevant patterns, but the high cost and computational complexity make it infeasible for consumer level applications \citep{lu2017}.

\citet{vangrondelle2017a} describe that the variation across the visible spectrum (400 to 700 nm) comes from the light-absorption properties of the pigments involved. The authors state that plant chlorophylls absorb strongly in the blue region (between 400 and 500 nm) and in the red region (around 650 to 680 nm). 
However, minimal absorption is observed in the green region (around 530 nm). This strong absorption in the blue and red wavelengths, along with weak absorption in the green, explains why chlorophyll appears green. As the amount of chlorophyll increases, visible reflectance decreases significantly, particularly 
within the blue and red region \citep{sharpe1972}. The paper by \citet{vangrondelle2017a} also mentions that other pigments, such as carotenoids, affect the overall spectral profile by absorbing light within the visible range. \citet{sharpe1972} also concluded that NIR reflectance is closely 
related to fruit firmness. Firmer fruits, such as apples and pears exhibit high solar NIR reflectance, typically between 53\% and 75\%. In contrast, softer fruits such as tomatoes and plums show much lower reflectance, usually ranging from 21\% to 37\%. The authors suggest that changes in spectral reflectance values in the 
NIR region largely result from internal structural changes. The high reflectance value in firm fruits comes from internal scattering of light caused by large intercellular air spaces.  As the fruit softens, the permeability of cell membrane increases, causing the intercellular spaces and cell wall micropores to fill with liquid, reducing the scattering of light. 
As a result, NIR light penetrates more deeply into the tissue, where it gets absorbed instead of being reflected. These studies make the correlation between fruit physiology and spectral intensity data in the visible-NIR range evident enough to be studied for non-invasive assessment of fruit ripening.

\subsection{\textbf{Proposed Solution}}
This study proposes a novel, multispectral approach for distinguishing mango and banana samples ripened using calcium carbide from safely ripened ones (consisting of
naturally ripened and ethylene ripened samples), while simultaneously
estimating the ripening progression (in \%) and predicting the
remaining shelf life (in days).The spectral profile of fruit samples ripened using all three ripening methods is obtained at different stages of ripening progression, using the AS7265x spectral triad sensor. While AS7265x has been validated for a number of
agricultural applications including estimating the percentage of grass cover and vine vigour \citep{ducanchez2022},  and evaluating the quality status
of intact olive fruits \citep{noguera2022}, its potential for detecting the use of chemical ripening agents in fruits remains underexplored.
To address this gap, the proposed study details the development of an XGBoost-based learning algorithm, utilizing a robust feature set derived from spectral intensity data across 18 discrete wavelengths in conjunction with environmental parameters. This methodology provides a robust framework for the classification of fruit samples on the basis of ripening method and estimating the remaining shelf life, in real time.
\\\\
% \\The remainder of this manuscript is structured as follows. Section 2 establishes the theoretical background for the study alongside a literature survey of existing methods for detecting the use of calcium carbide in fruit ripening. The materials and methods used, including the experimental design and analytical workflow are detailed in section 3. This includes the data acquisition process, defining the output labels, and detailing the steps invovled in computational procedures such as data augmentation, PCA, Spearman correlation and the XGBoost based architecture used for training the learning algorithm. The results and analysis are covered in section 4, which includes the spectral progression and its physiological interpretation,
% feature extraction, handling high dimensionality of spectral data, environmental calibration, and comprehensive model benchmarking. At the end, section 5 summarizes the findings and concludes the study.
The remainder of this manuscript is structured as follows. The materials and methods used, including the experimental design and analytical workflow are detailed in section 2. This includes the data acquisition process, defining the output labels, and detailing the computational procedures such as data augmentation, PCA, Spearman's rank correlation and the XGBoost based architecture used for training the learning algorithm. The results and analysis are covered in section 3, which includes the spectral progression and its physiological interpretation,
feature extraction, handling high dimensionality of spectral data, environmental calibration, and comprehensive model benchmarking. At the end, section 4 summarizes the findings and concludes the study.

%%%%%%%%%%%%%%%%%%%%%%%%%%%%%%%%%%%%%%%%%%%%%%%%%%%%%%%%%%%%%%
\section{Materials and Methods}

The experimental framework for the proposed study consists of three key stages: (i) spectral data acquisition using three distinct ripening methods, (ii) rigorous data visualization and analysis for feature extraction, and (iii) training a learning algorithm for ripening method classification and predicting the shelf life analytics. \
The following subsections detail the dataset creation process, output labels for learning algorithm, hardware setup, and the analytical tools used.

\subsection{\textbf{Dataset Creation}}
\begin{figure}[htbp]
    \centering
    \begin{subfigure}[b]{0.48\textwidth}
        \centering
        \includegraphics[width=5cm]{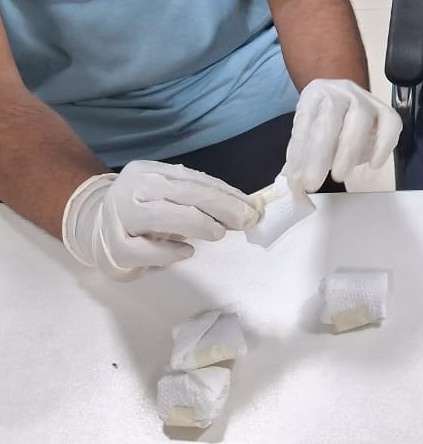} 
        
        \caption{Calcium Carbide sachets}
        \label{fig:image1}
    \end{subfigure}%
    \vspace{0.2cm}
    \begin{subfigure}[b]{0.48\textwidth}
        \centering
        \includegraphics[width=5cm]{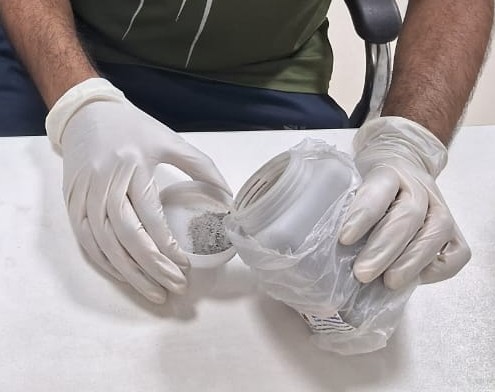}
        \caption{Calcium Carbide powder}
        \label{fig:image2}
    \end{subfigure}%
    \caption{Preparation of sachets for artificial ripening where CaC\textsubscript{2} powder is wrapped in tissue paper prior to placement with fruit samples.}
    \label{fig:combined_figures}
\end{figure}

A total of 100 raw fruit samples, comprising 60 bananas and 40 mangoes, are sourced from Azadpur Mandi in Delhi through a vendor on the campus of the National Institute of Technology, Delhi. These fruits are divided into three batches to be ripened under different conditions. The first batch consists of acetylene-ripened fruits, where 5 g of CaC\textsubscript{2} powder, wrapped in tissue paper (Figure~\ref{fig:combined_figures}), is placed with each kilogram of fruit samples, 
maintaining the concentration of CaC\textsubscript{2} at 5g/kg for both banana and mango samples \citep{adeyemi2018,essien2018}. The second batch is ripened using FSSAI-approved ethephon sachets \citep{fssai2020}. One sachet is used per 5 kg of fruits in accordance with the specification on product packaging. The third batch is naturally ripened using dry straw grass. Each batch is further divided into two groups based on storage 
conditions. One group is kept in an air-conditioned room at a temperature ranging from 20°C to 27°C and another group is stored outdoors at a temperature between 30°C and 33 °C. Observation concluded upon the spoilage of all fruit samples within an 11-day period, yielding a dataset of 1,172 unique temporal data points for subsequent analysis. These are longitudinal measurements collected from the same 100 individual fruits over a period of 11 days.

\subsection{\textbf{Color and Firmness Stage}}
The color and firmness of the fruit are used to determine the ripeness percentage. The change in peel color is recorded on a scale from 0 to 7, and firmness is rated on a scale from 0 to 4, depending on the fruit's softness. Analogous to the Von Loesecke ripening scale \citep{gomes2013}, the color scale is defined as follows: S0 - Green, S1 - Yellow color break, S2 - Mostly green with yellow texture, S3 - Mixed yellow and green, S4 - Yellow dominant, S5 - Almost full yellow, S6 - Bright yellow, and S7 - Blackish color. The firmness scale is defined as: S0 - Fully Firm, S1 - Soft patches in some area, S2 - Mixed firmness, S3 - Almost the entire body is soft, S4 - Fully Soft.
\begin{figure*}[t!]
    \centering
    \begin{subfigure}[b]{0.24\textwidth}
        \centering
        \includegraphics[height=4cm, width=\textwidth, keepaspectratio]{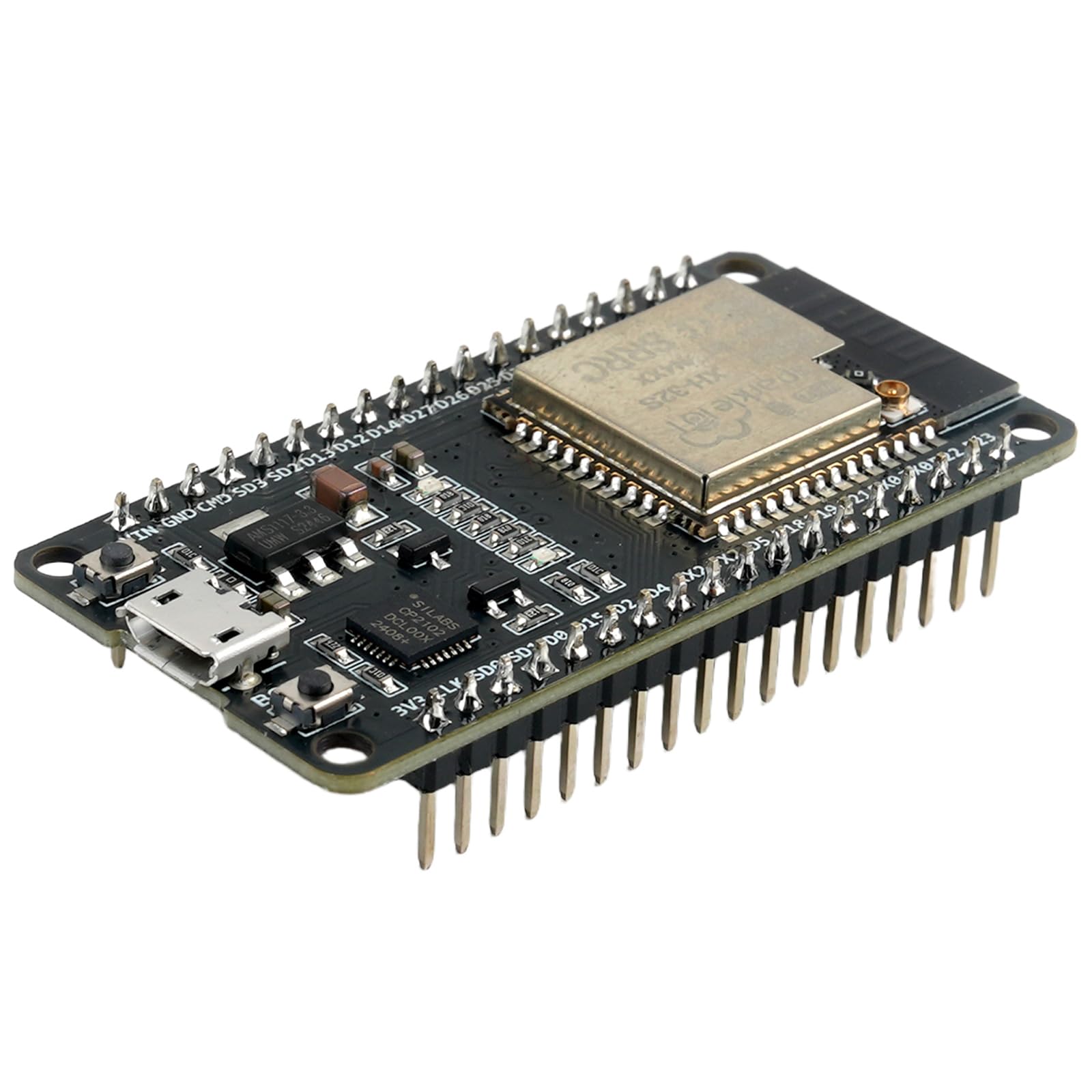} 
        \caption{ESP32 Module}
        \label{fig:esp}
    \end{subfigure}% <--- Percent signs prevent unwanted line breaks
    \vspace{0.2cm}
    \begin{subfigure}[b]{0.24\textwidth}
        \centering
        \includegraphics[height=4cm, width=\textwidth, keepaspectratio]{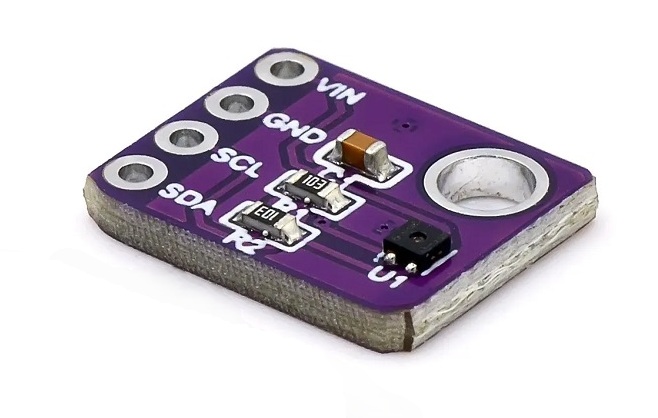}
        \caption{SHT40 Sensor}
        \label{fig:sht}
    \end{subfigure}%
    \hfill
    \begin{subfigure}[b]{0.24\textwidth}
        \centering
        \includegraphics[height=4cm, width=\textwidth, keepaspectratio]{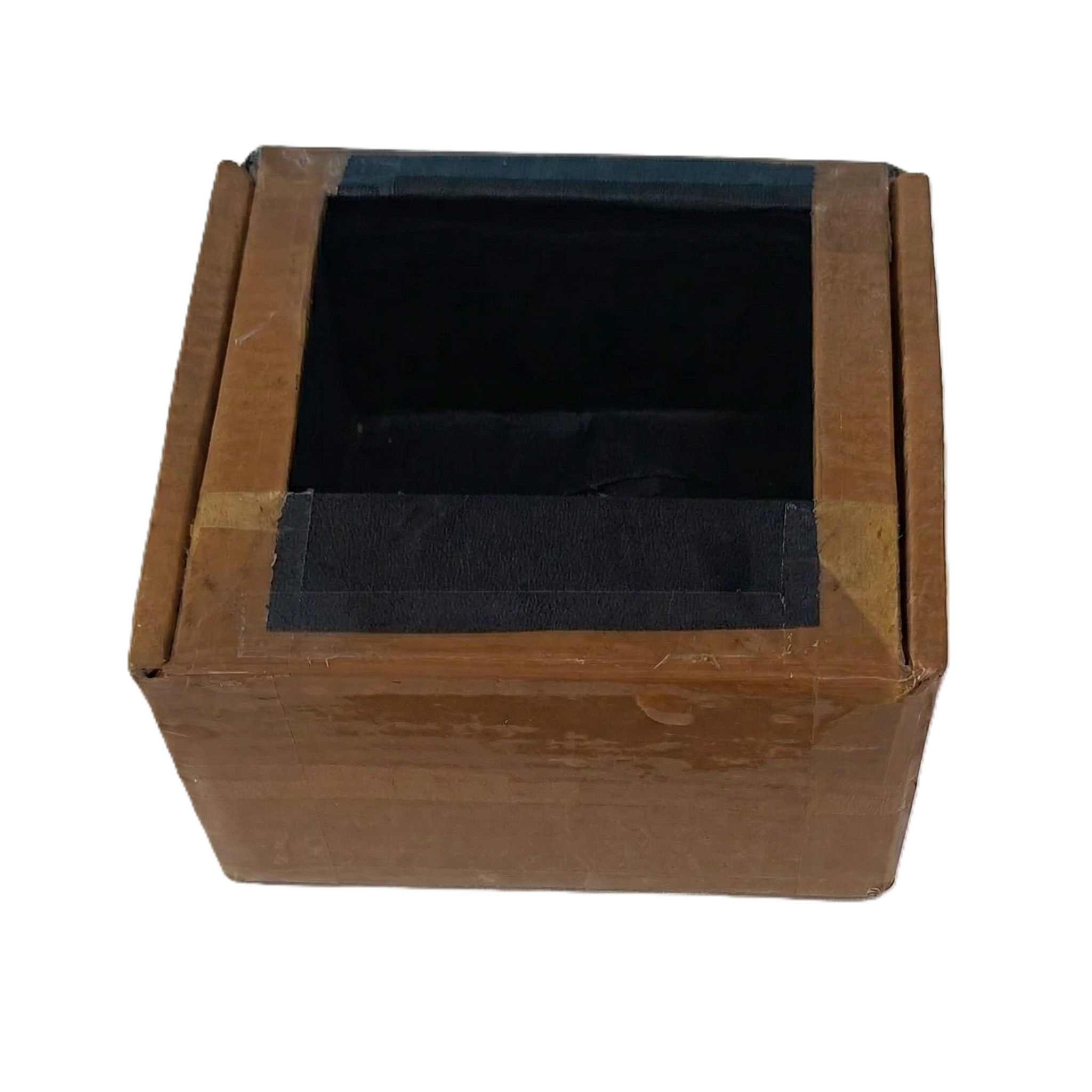}
        \caption{Enclosed casing}
        \label{fig:box}
    \end{subfigure}%
    \vspace{0.2cm}
    \begin{subfigure}[b]{0.24\textwidth}
        \centering
        \includegraphics[height=4cm, width=\textwidth, keepaspectratio]{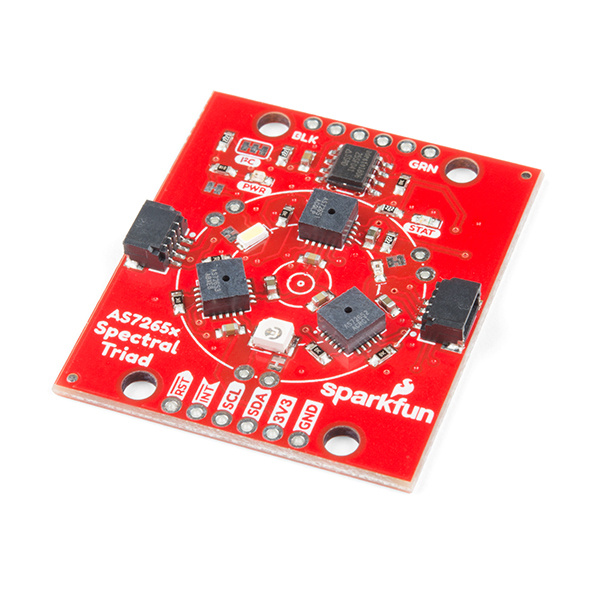}
        \caption{AS7265x Sensor}
        \label{fig:as7265x}
    \end{subfigure}%   
    \caption{Hardware components used in the experimental setup consisting of (a) ESP32 microcontroller, (b)SHT40 temperature and humidity sensor, (c) 10 cm × 10 cm × 6 cm enclosed  casing, coated with black felt sheet, (d) and AS7265x spectral triad sensor.}
    \label{fig:hardware_components}
\end{figure*}

\subsection{\textbf{Ripeness Percentage and Shelf Life}}

Ripeness percentage is used in this study as a heuristic index designed to provide a numerical description of ripening progression, rather than a direct biochemical quantification \citep{abbott1999}. For mangoes (Mangifera indica), the ripeness percentage is based on the fact that the studied cultivar stays predominantly green throughout the ripening process, while its firmness changes. A higher weight is assigned to firmness as compared to color to take into account this physiological behavior \citep{jha2010}. Firmness and color are combined using empirically selected weightings (80\% to firmness and 20\% to color). In contrast, bananas (Musa acuminata) exhibit a noticeable color transition during the entire course of ripening, making surface color a reliable indicator of maturity. Therefore, a higher weight is assigned to color as compared to firmness (60\% to color and 40\% to firmness) \citep{dadzie1997}. This index for calculating ripeness percentage is only intended as a relative progression metric for temporal modeling and supervised learning; not as a universal standard. A 5\% baseline offset is included to set a realistic starting point, considering pre-harvest physiological maturity \citep{blasco2003}.\\
To determine shelf life, every fruit gets a unique serial number, which is tracked throughout the study. The last recorded day that a fruit’s serial number appears in the dataset is considered its true observed lifespan. The observation for an individual fruit is terminated when the fruit reaches spoilage condition, after which no further measurements are included for that fruit. If $D$ represents the last observed day and $d$ is the current day (the number of days since the observation of ripening started), then the remaining shelf life of the fruit on any day is calculated as $(D-d)$ days.

\subsection{\textbf{Hardware Setup}}

\begin{figure}[H]
    \centering
    % First Image
    \begin{subfigure}[b]{0.48\textwidth}
        \centering
        \includegraphics[width=\textwidth]{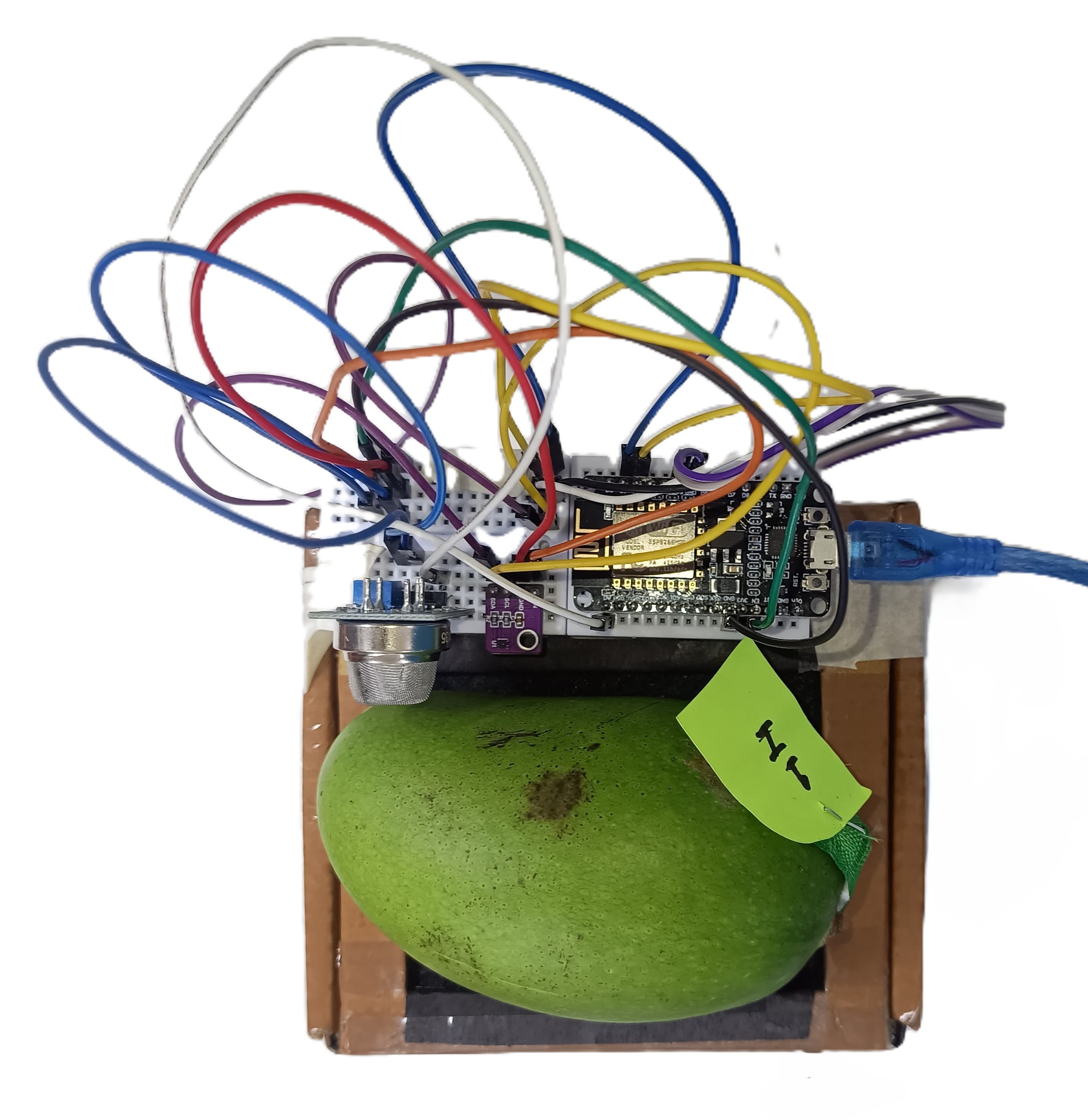}
        \caption{The hardware setup actually used in the study}
        \label{fig:original_photo}
    \end{subfigure}%
    \vspace{0.2cm}
    \begin{subfigure}[b]{0.48\textwidth}
        \centering
        \includegraphics[width=\textwidth]{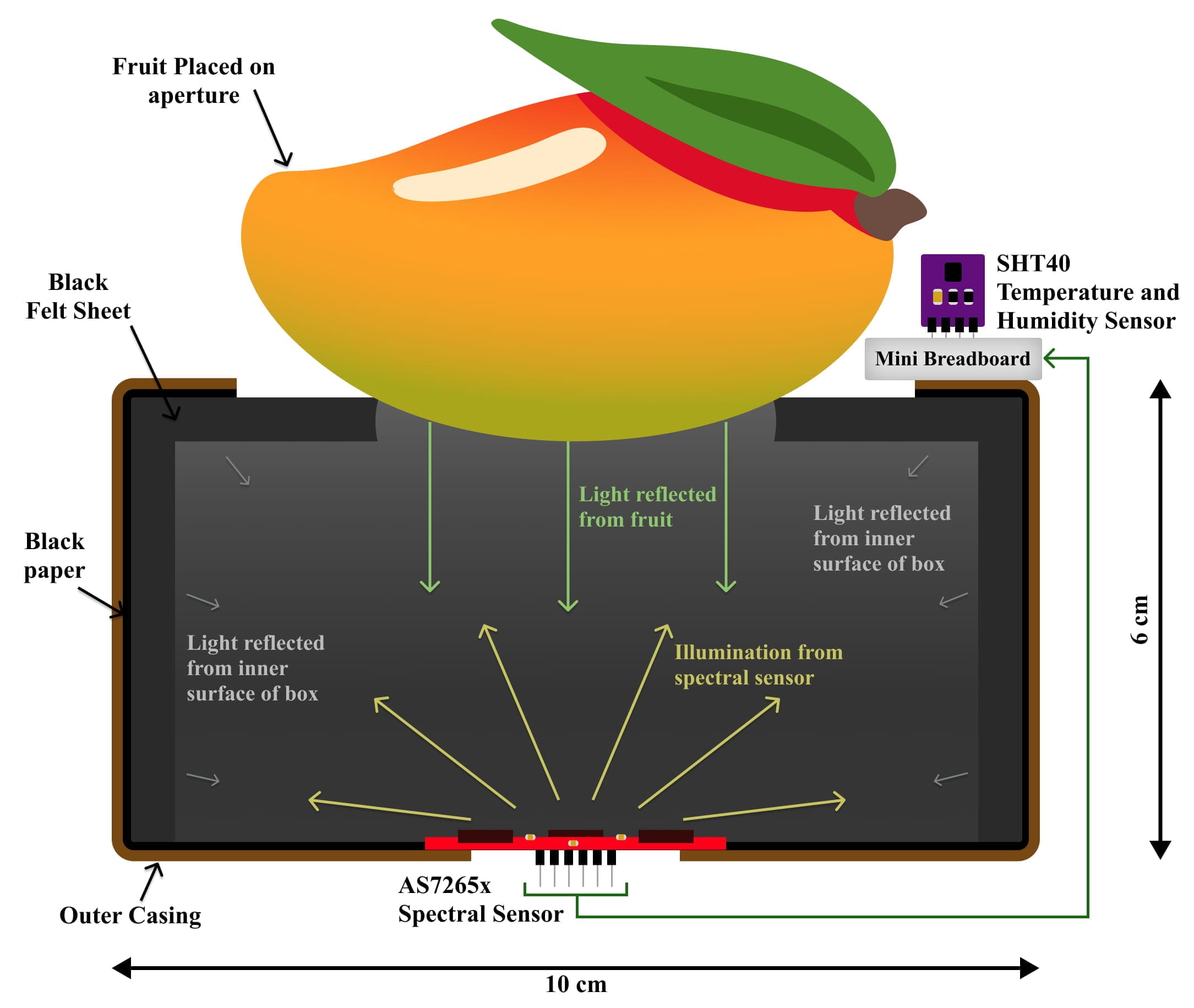}
        \caption{Schematic diagram detailing the internal structure of hardware setup}
        \label{fig:schematic}
    \end{subfigure}%
    \caption{Hardware configuration and spectral sensing mechanism: (a) Physical prototype utilized for the experimental study, and (b) Cross-sectional schematic detailing the optical path and internal geometry. The enclosed sensing chamber is lined with a matte black felt sheet to suppress internal scattering and stray light, ensuring that the intensity data exclusively represents the spectral profile of the fruit.}
    \label{fig:combined_setup}
\end{figure}

\begin{figure}[H]
    \centering
        \includegraphics[width=0.48\textwidth, keepaspectratio]{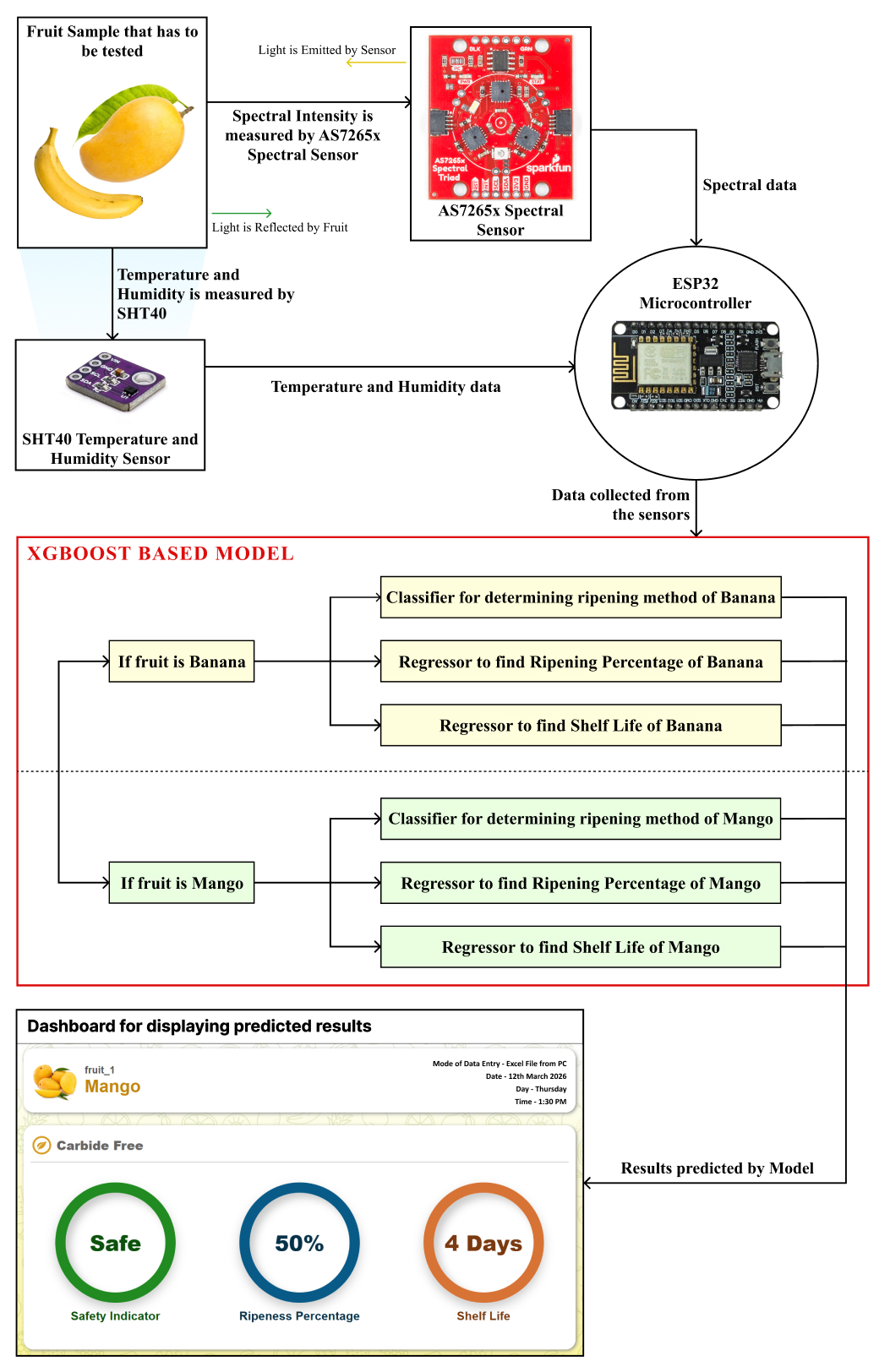}
        \caption{Comprehensive workflow of the proposed non-invasive ripening assessment system. The architecture integrates the AS7265x spectral triad sensor and  SHT40 temperature and humidity sensor via an ESP32 microcontroller. Data is processed through an XGBoost-based learning algorithm consisting of fruit-specific classifiers for ripening method detection and regressors for estimating ripening percentage and remaining shelf life, with final outputs visualized on a real time monitoring dashboard.}    
        \label{fig:workflow}
    \end{figure}

Spectral intensity data of the fruit samples is acquired using a custom-built hardware setup (Figure~\ref{fig:combined_setup}) that integrates an AS7265x multispectral sensor, an SHT40 temperature and humidity sensor, and an ESP32 microcontroller (Figure~\ref{fig:hardware_components}).
The setup is housed in a 10 cm × 10 cm × 6 cm cuboidal casing. The AS7265x spectral sensor is mounted on the bottom face (10 cm $\times$ 10 cm) of the casing, while the opposite top face features an aperture sized to accommodate a portion of the fruit's surface. This setup uses the inbuilt LEDs present in spectral sensor for illumination and it is ensured that there is a distance of 5 cm between sensor and fruit surface. The inner surface of box is coated with black felt sheet to suppress secondary internal reflections and stray light. The ESP32 microcontroller and SHT40 sensor are mounted on a breadboard at outer surface of the casing and connections with the microcontroller are made using jumper wires.
The ESP32 microcontroller is integrated with Google Sheets using Google Apps Script to automatically log the spectral data being collected through the microcontroller into the spreadsheet. Spectral data for each day is stored in a separate spreadsheet and each sheet includes  serial number, fruit type, fruit's unique sample ID, ripening method, day since observation started, raw spectral readings from all eighteen channels, temperature and humidity readings from the SHT40 sensor, color stage, firmness stage, calculated ripening percentage, and comments on the condition of each fruit sample, as the columns. The results obtained by testing the fruit samples are displayed on a dashboard, as shown in Figure~\ref{fig:workflow}.

\subsection{\textbf{Data Preprocessing and Feature Engineering}}

In the proposed study, noise with a mean $(\mu)$ of 1.0 and a standard deviation $(\sigma)$ of 0.02 is injected to introduce small variations in the training set. Sampling from the distribution $\mathcal{N}(1, 0.02^2)$ ensures that overall scale of the data does not change. Furthermore, the multiplicative noise scales with the magnitude of the data, such that value changes less if original value is smaller and large original values experience greater absolute change. 
This enhances model stability as addition of these controlled disturbances exposes the model to realistic variations that might occur while taking the readings.
Further, PCA is employed for reducing the dimensionality of the 18 wavelength spectral data into principal components that capture the maximum variance \citep{ding2018}.
One-Way Analysis of Variance (ANOVA) is performed independently for each ripening method to evaluate the impact of environmental conditions like temperature and relative humidity on spectral readings. In each analysis, the whole set of samples exposed to a specific ripening treatment serves as the population. This population is then divided into two groups namely "Indoor" and "Outdoor" based on environmental condition with different temperature and relative humidity. The grouping of samples based on environmental condition acts as the independent categorical variable, while the spectral intensity measured at each wavelength (410nm - 940nm) acts as continuous dependent variable. The p-value or observed significance level for each wavelength is used to determine whether spectral intensity differs significantly between the indoor and outdoor environmental conditions.
\\\\
In order to establish a robust set of input features, all possible ratios of raw spectral intensity values at the 18 wavelengths are taken into consideration, and their progression over time is analyzed. It is critical in differentiating between ripening methods and estimation of shelf life, as spectral ratios effectively amplify relative physiological changes \citep{thenkabail2000}. The raw spectral intensity values are standardized using their respective z-scores, which ensures that the effect of outliers is minimized. For each fruit type, the mean and standard deviation are calculated using all training observations across the available days and ripening methods, and the resulting parameters are used to standardize the corresponding training, validation, and hold-out test observations.
The potential of each feature as a discriminator between ripening methods or indicator of ripening progression is evaluated using two primary metrics. The IoU of IQRs is used to quantify overlap across ripening methods, and the Spearman's rank correlation coefficient is used to assess the monotonic progression of the features across the three ripening methods. 
In this way, all the possible input features are identified for each predictive modelling task. These features also include principal components covering variance of spectral intensity across all 18 wavelengths, along with temperature and humidity as environmental parameters.
Consequently, to reduce the complexity of model by removing non-contributing features and understand how each input feature is contributing to model performance, SHAP is applied to select the most relevant subset of input features out of all the possible features explored.

\subsection{\textbf{Machine Learning Framework}}

In the proposed study, an XGBoost based machine learning framework is developed for classification of ripening method used while simultaneously estimating the shelf life and ripeness percentage. A dedicated set of three models comprising of one classifier and two regressors is constructed independently for both banana and mango samples. All models are trained using the data acquired from the aforementioned hardware setup and have their own set of optimized input features, validated using SHAP analysis. To prevent data leakage caused by the temporal nature of the dataset, data is split into 80\% training and 20\% hold-out test sets using a Group Shuffle Split based on the fruit's unique serial numbers. This ensures that all temporal observations of a specific fruit remain strictly within either the training or the test set. Furthermore, augmentation (noise injection), SMOTE, and all data-dependent preprocessing and feature-selection operations are performed exclusively within the training partition. Model stability is rigorously evaluated using a 5-fold Group K-Fold cross-validation strategy on the training partition. The final reported performance metrics are computed exclusively on the unaugmented hold-out test set. For classification, the positive class is calcium carbide-ripened fruits and the negative class is safely ripened fruit. The final XGBoost hyperparameters selected by GridSearchCV are task and fruit-specific, including \texttt{max\_depth}, \texttt{learning\_rate}, \texttt{n\_estimators}, \texttt{subsample}, \texttt{colsample\_bytree}, and \texttt{gamma}.
For the ripening method classification task, SMOTE is applied only to the training set within the pipeline, which allows the classifier to learn from balanced training data while ensuring that oversampling does not leak information into the test set. The hold-out test set is not used during preprocessing, feature selection, hyperparameter tuning, data augmentation, or SMOTE.

\section{Results and Analysis}
This section presents a comprehensive evaluation of the proposed non-invasive fruit assessment system, and discusses the results obtained. The analysis is structured into six parts. First, the temporal progression of spectral intensity across different ripening methods is examined, 
correlating variations in spectral profile to underlying physiological changes in the fruit. Second, statistical techniques, including Spearman's rank correlation coefficient and Intersection over Union (IoU) metrics, are applied to extract and validate the most discriminative features from the raw spectral data. Third, use of PCA in handling the high dimensional spectral data is studied. Fourth, the impact of environmental parameters, including temperature and humidity on spectral progression is studied. Fifth, the proposed XGBoost framework and feature sets used are detailed and finally, a thorough benchmarking is performed to analyze the predictive capabilities of various machine learning architectures in both classifying the ripening methods and quantifying the remaining shelf life and ripeness percentage.

\begin{figure}[h]
    \centering
    % First Image
    \begin{subfigure}[b]{0.48\textwidth}
        \centering
        \includegraphics[width=\textwidth]{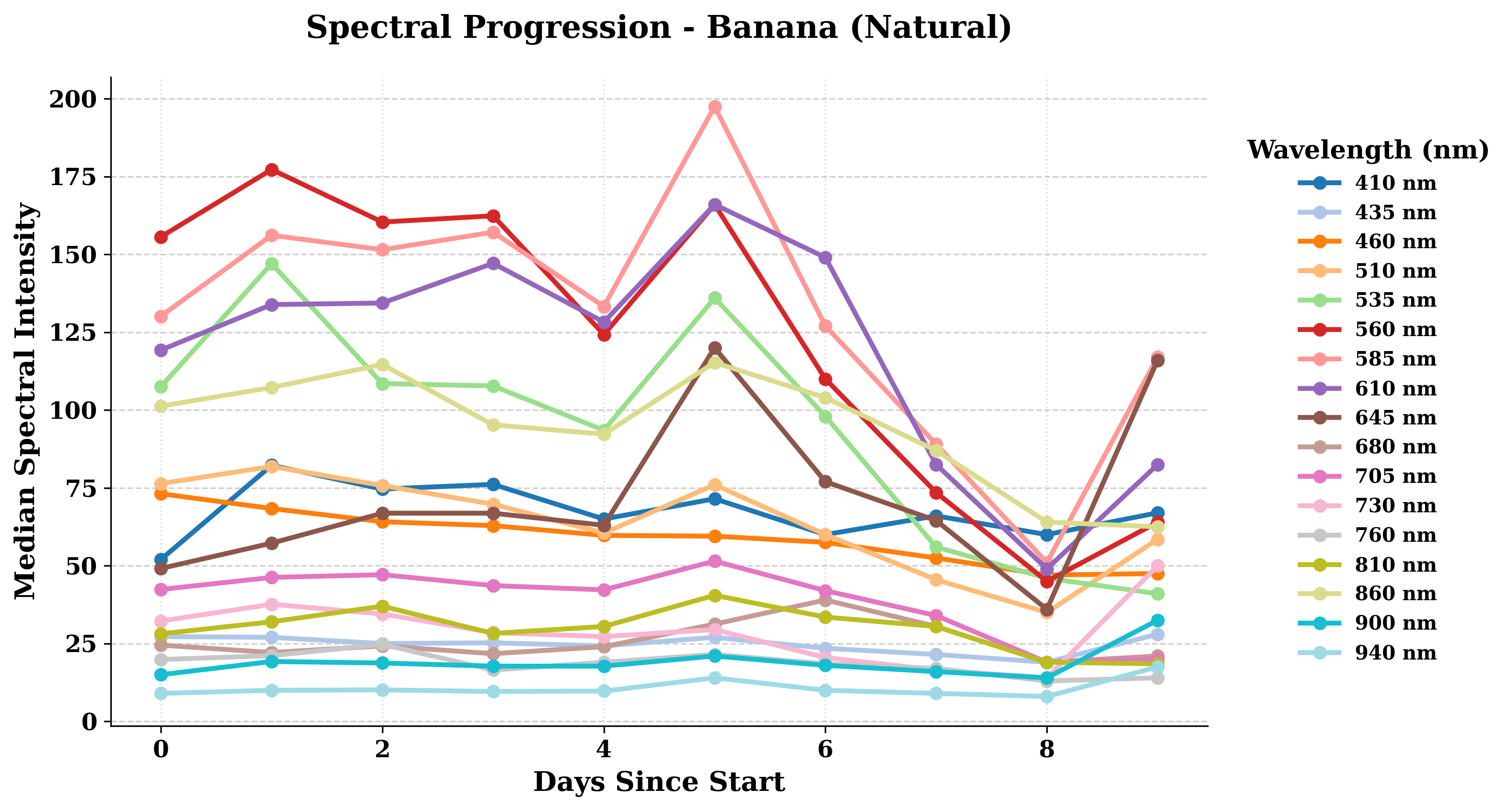}
        \caption{}      
    \end{subfigure}%
    \vspace{0.2cm}
    % Second Image
    \begin{subfigure}[b]{0.48\textwidth}
        \centering
        \includegraphics[width=\textwidth]{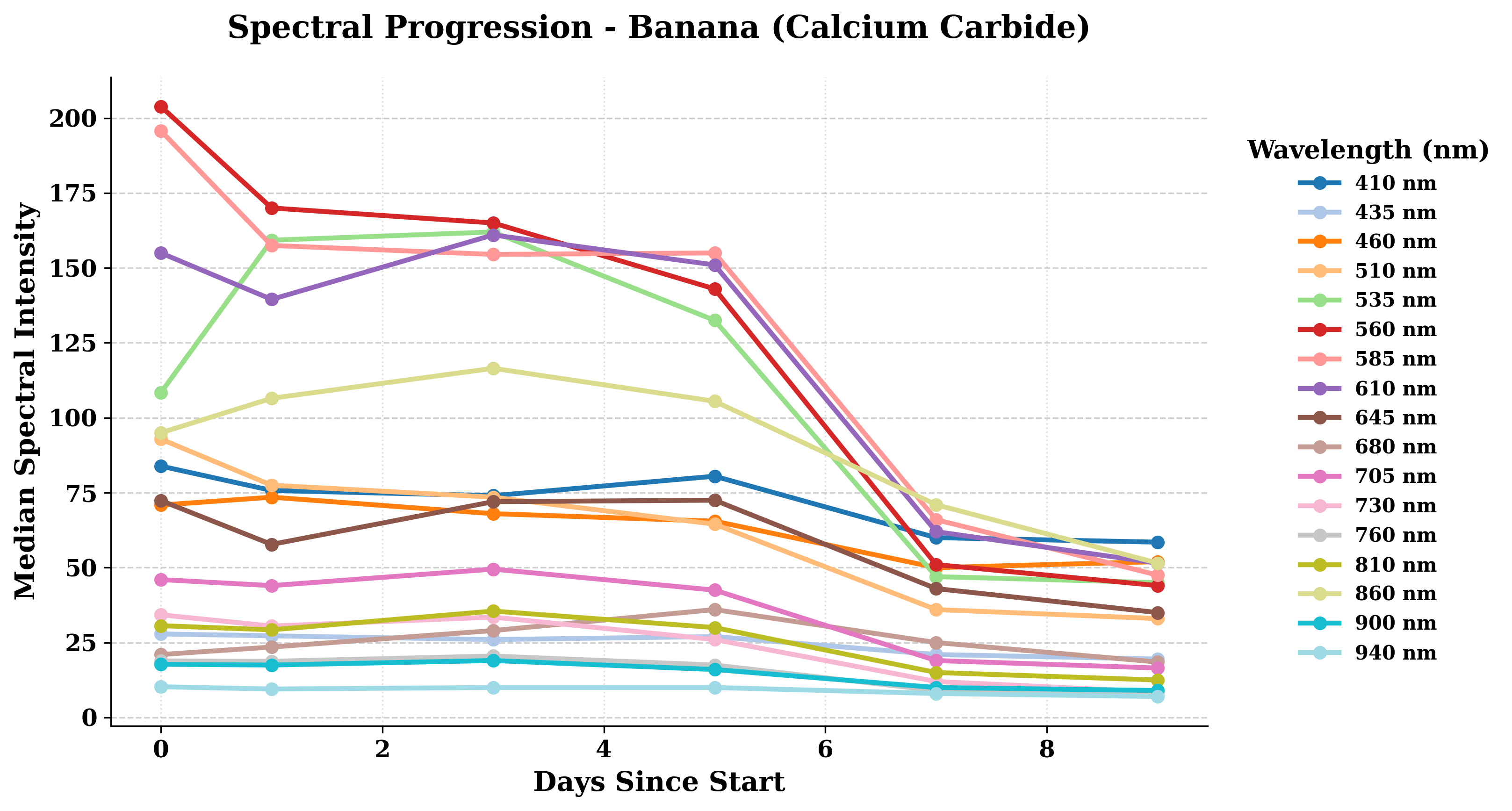}
        \caption{}\end{subfigure}%
    \caption{Temporal progression of median spectral intensity for banana samples under (a) natural ripening and (b) calcium carbide-induced ripening conditions. The data represents 18 distinct wavelengths ranging from 410 nm to 940 nm. Note the sharper decline in intensity across visible and NIR bands in carbide-treated samples post Day 5, which is indicative of accelerated chlorophyll degradation and moisture loss compared to the natural ripening progression.} 
    \label{fig:wavelength_progression}
\end{figure}

\subsection{\textbf{Spectral Progression and Physiological Interpretation}}
The median spectral intensity profiles of Mangifera indica (mango) and Musa acuminata (banana) across 18 wavelengths, ranging from 410 nm to 940 nm are analyzed to study ripening progression under distinct ripening methods and environmental conditions. The median is preferred over the mean as a measure of central tendency to alleviate the influence of outliers and occasional sensor anomalies \citep{leys2013}. A consistent trend is obtained where the green-yellow range of visible region (535 nm–585 nm) and the “red edge” or early NIR region (610 nm–705 nm) exhibit pronounced changes during ripening, while spectral intensity at other wavelengths remains comparatively stable. This aligns with the color transformation of fruits \citep{vangrondelle2017a}. Chlorophyll degradation reveals underlying carotenoids, driving the macroscopic color transition from green to yellow.
\begin{figure}[H]
    \centering
    % First Image
    \begin{subfigure}[b]{0.45\textwidth}
        \centering
        \includegraphics[width=\textwidth]{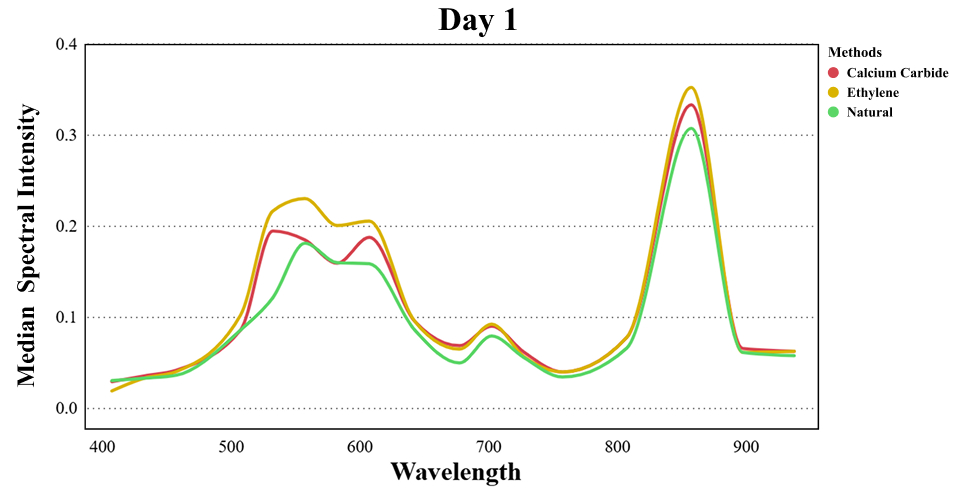}
        \caption{Spectral pattern at day-1}
        \label{fig:day_1}
    \end{subfigure}%
    \vspace{0.2cm}
    % Second Image
    \begin{subfigure}[b]{0.45\textwidth}
        \centering
        \includegraphics[width=\textwidth]{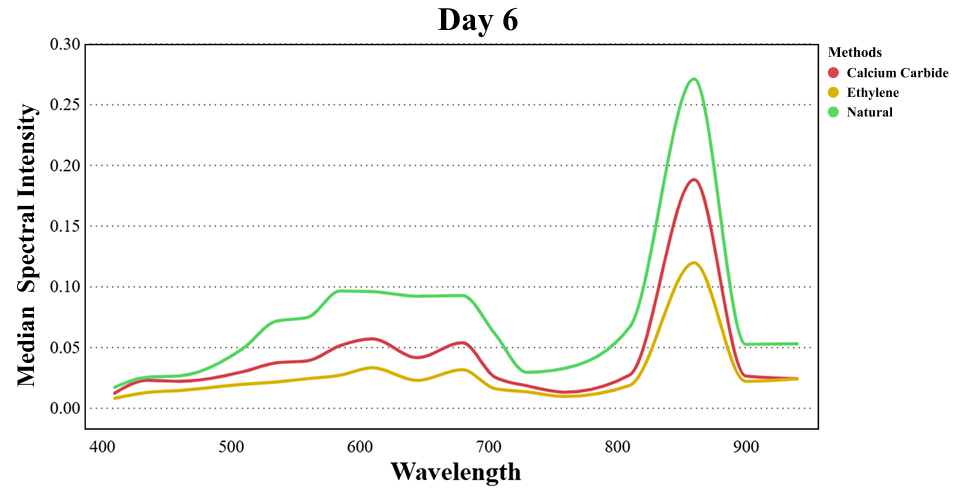}
        \caption{Spectral pattern at day-6}
        \label{fig:day_6}
    \end{subfigure}%
    \vspace{0.2cm}
    % Third Image
    \begin{subfigure}[b]{0.45\textwidth}
        \centering
        \includegraphics[width=\textwidth]{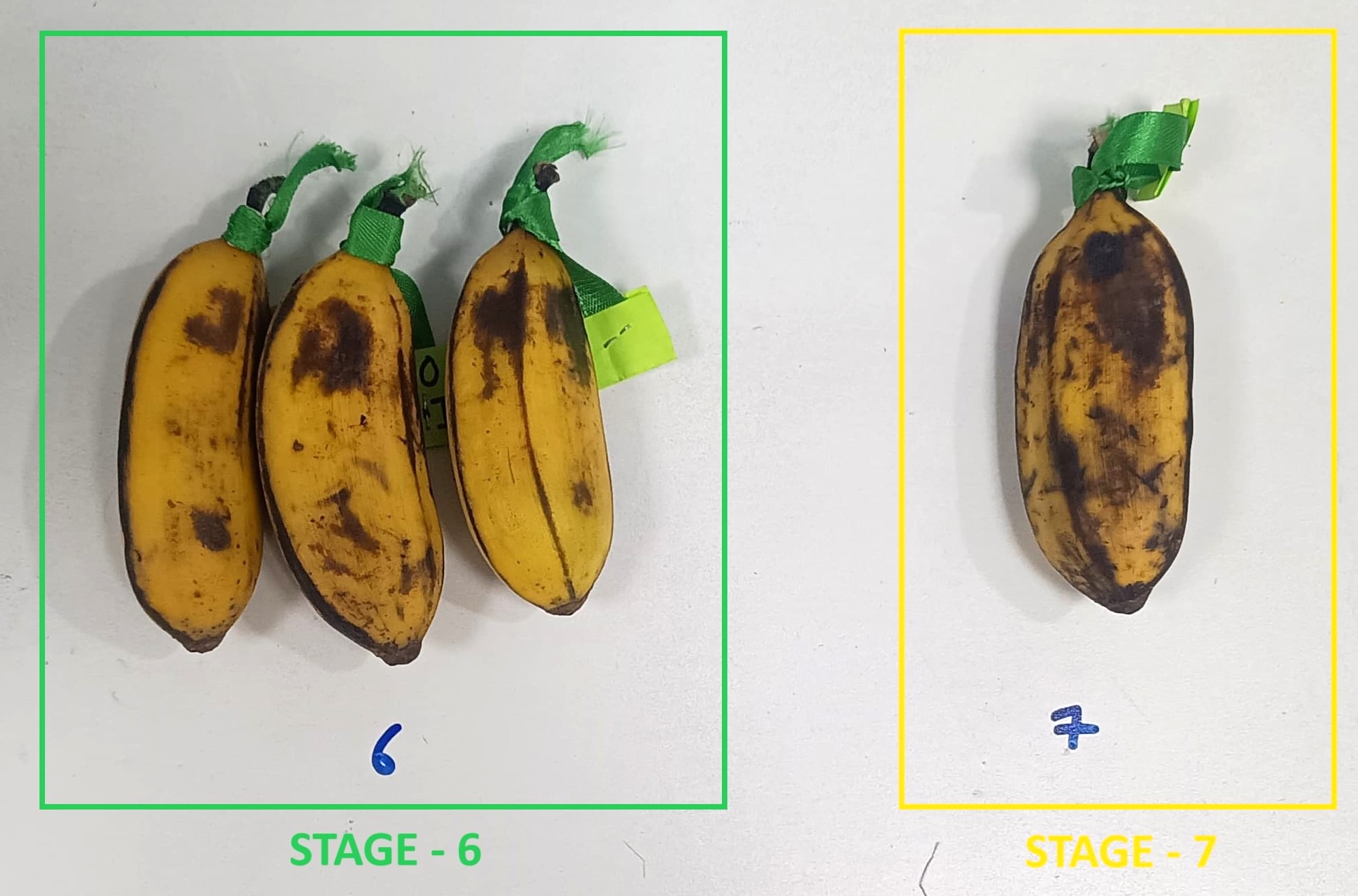}
        \caption{Naturally ripened Banana at day-6}
        \label{fig:banana_day_6}
    \end{subfigure}%  
    \caption{As the fruit ripens, the amount of chlorophyll decreases and the internal structure degrades. In case of banana, the overall spectral intensity decreases on day 6 (b) as compared to day 1 (a). The spectral intensity in visible region decreases because black spots begin to form on the banana peel (c), while the intensity in NIR region decreases due to degradation of cells.}
    \label{fig:chlorophyll_banana}
\end{figure}
As seen in Figure~\ref{fig:wavelength_progression}, calcium carbide-ripened bananas exhibit sharper intensity drops in the visible region after mid ripening stages (day 4) as compared to naturally ripened samples, indicating accelerated color transformation, which is not accompanied by a proportionate change in longer wavelengths, within the NIR region (705 nm–900 nm).
Across different ripening methods, near-infrared bands (760 nm–900 nm) show sensitivity to internal biochemical changes of the fruit rather than just its surface color. Variations in these bands indicate moisture loss, starch to sugar conversion, and loss of firmness. \citep{pourdarbani2021}. This makes them useful for non-invasive assessment of fruit ripening.
Hence, the combined spectral profile of fruit samples in visible and NIR regions captures both external and internal changes that take place as the fruit matures. This behaviour is consistent with the characterization of calcium carbide-induced ripening as predominantly “cosmetic” \citep{deshi2024}, where the external appearance of fruit advances faster than internal growth. Using the spectral progression of multiple wavelengths simultaneously makes quantitative modelling challenging, making further statistical analysis crucial, in order to identify the most informative features. The spectral intensity in visible spectrum decreases as the amount of chlorophyll increases \citep{vangrondelle2017a,sharpe1972}, therefore the less intensity of visible light in ethylene-ripened and carbide-ripened bananas as compared to naturally-ripened samples (Figure~\ref{fig:chlorophyll_banana}) is consistent with the observation made by \citet{maduwanthi2019} that artificially ripened banana has higher levels of chlorophyll at the bright yellow color stage (stage 6) than the naturally ripened banana.

\subsection{\textbf{Feature Extraction and Statistical Analysis}}
\begin{figure}[H]
    \centering
    % --- Image 1 ---
    \begin{subfigure}[b]{1.0\linewidth}
        \centering
        \includegraphics[width=\linewidth, height=4.6cm, keepaspectratio]{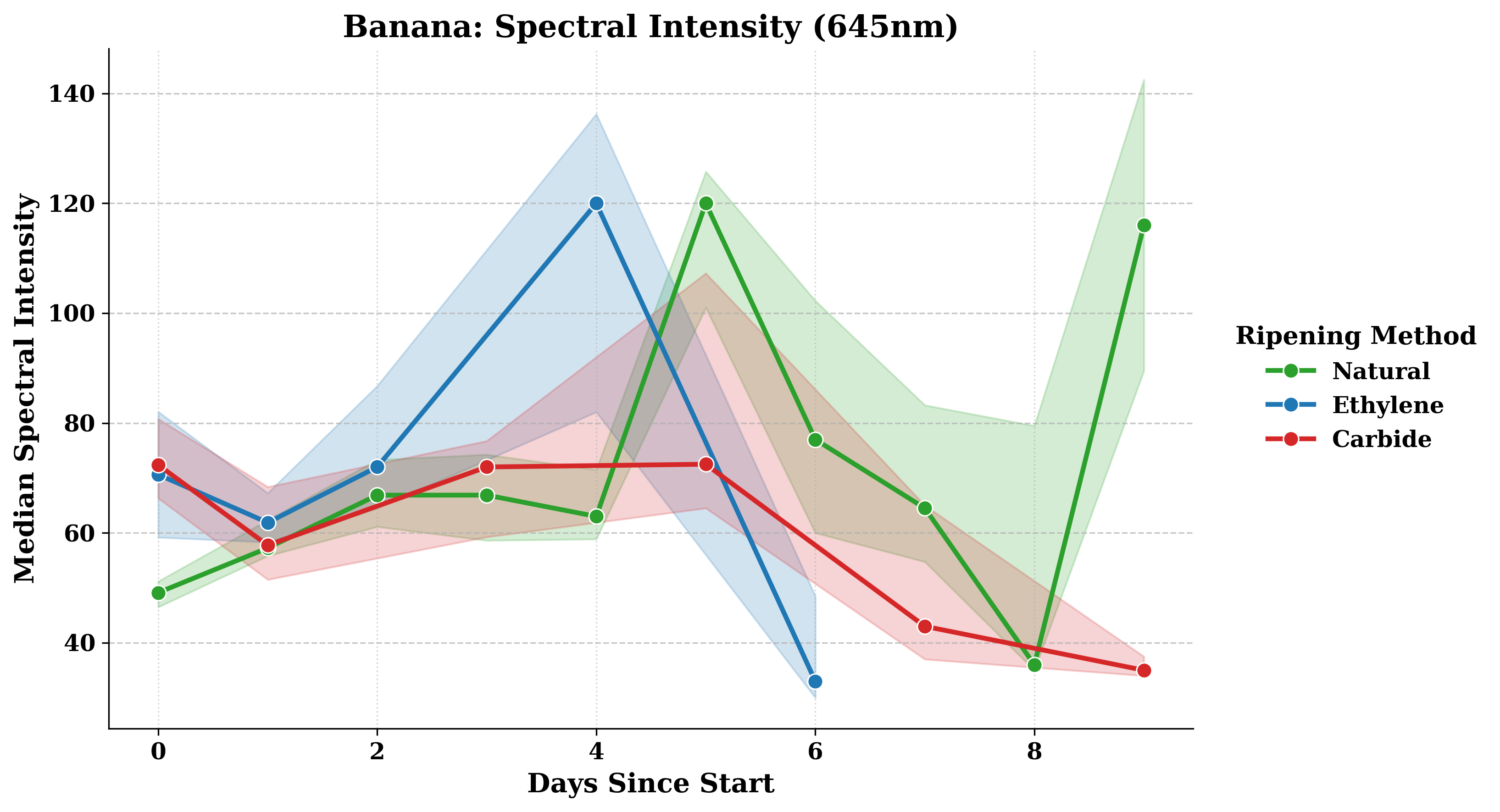} 
        \caption{}
        \label{fig:banana1}
    \end{subfigure}%
    \vspace{0.2cm}
    % --- Image 2 ---
    \begin{subfigure}[b]{1.0\linewidth}
        \centering
        \includegraphics[width=\linewidth, height=4.5cm, keepaspectratio]{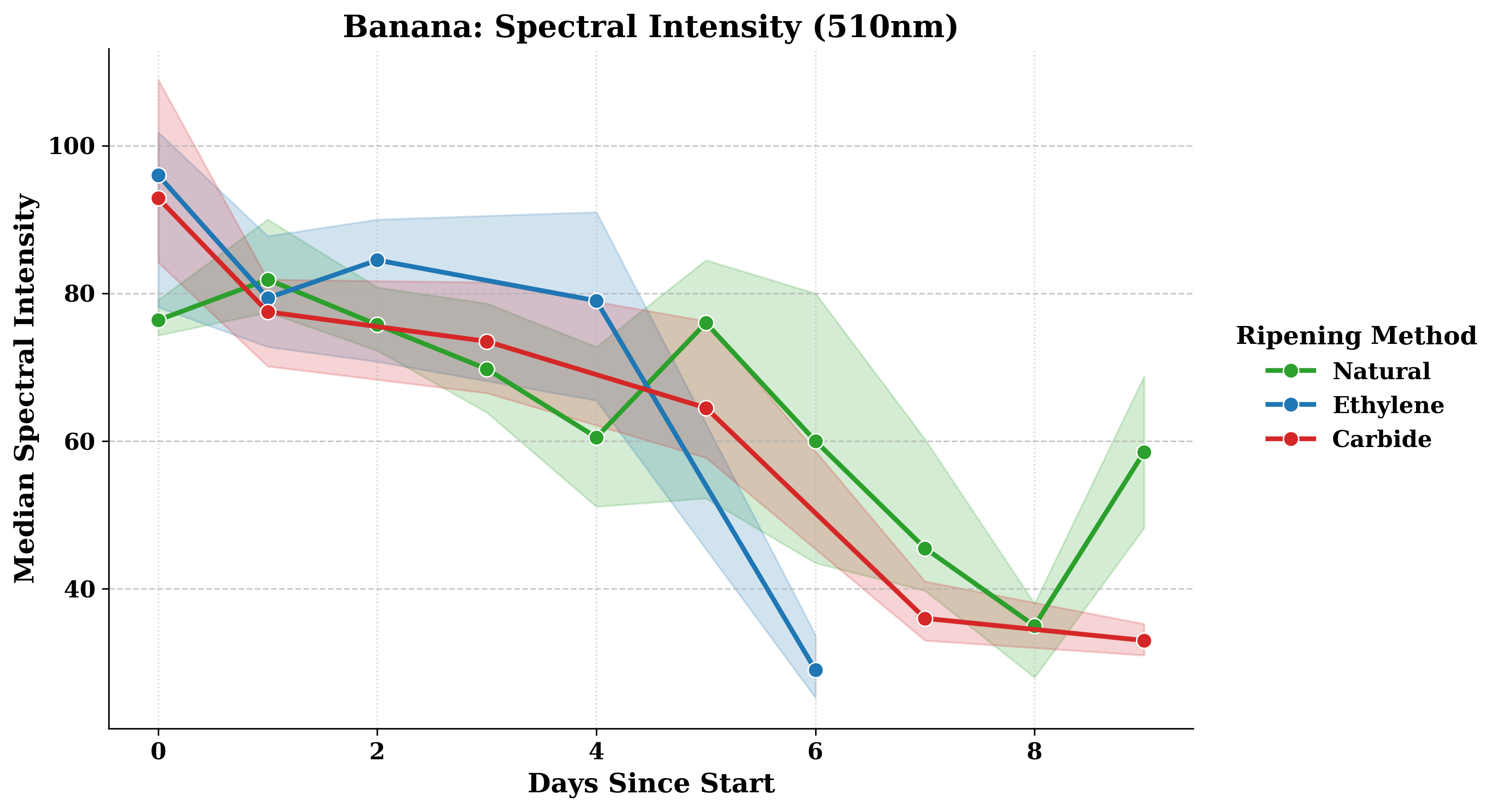}
        \caption{}
        \label{fig:banana2}
    \end{subfigure}%
    \vspace{0.2cm}
    % --- Image 3 ---
    \begin{subfigure}[b]{1.0\linewidth}
        \centering
        \includegraphics[width=\linewidth, height=4.5cm, keepaspectratio]{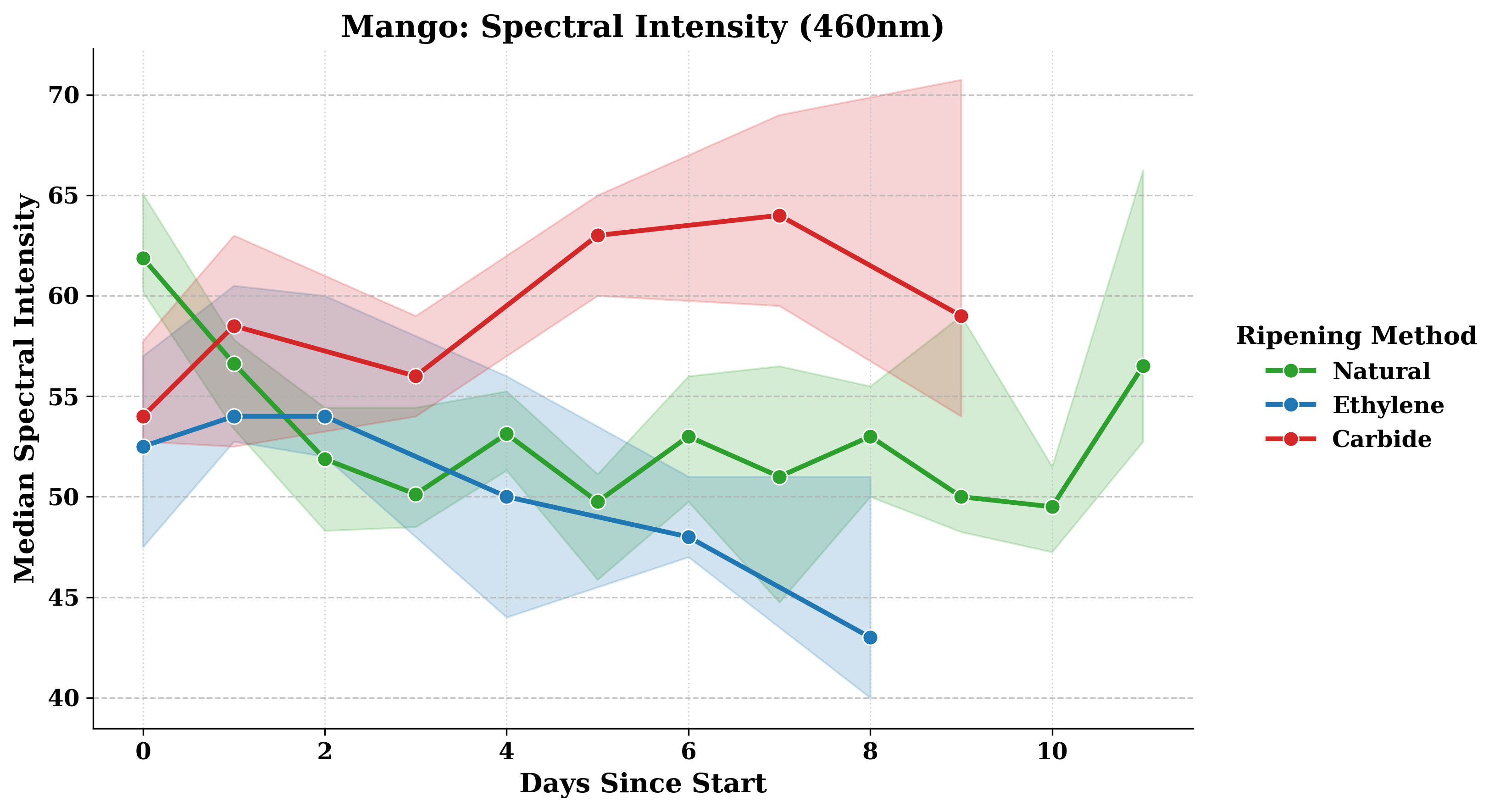}
        \caption{}
        \label{fig:mango1}
    \end{subfigure}%
    \vspace{0.2cm}
    % --- Image 4 ---
    \begin{subfigure}[b]{1.0\linewidth}
        \centering
        \includegraphics[width=\linewidth, height=4.5cm, keepaspectratio]{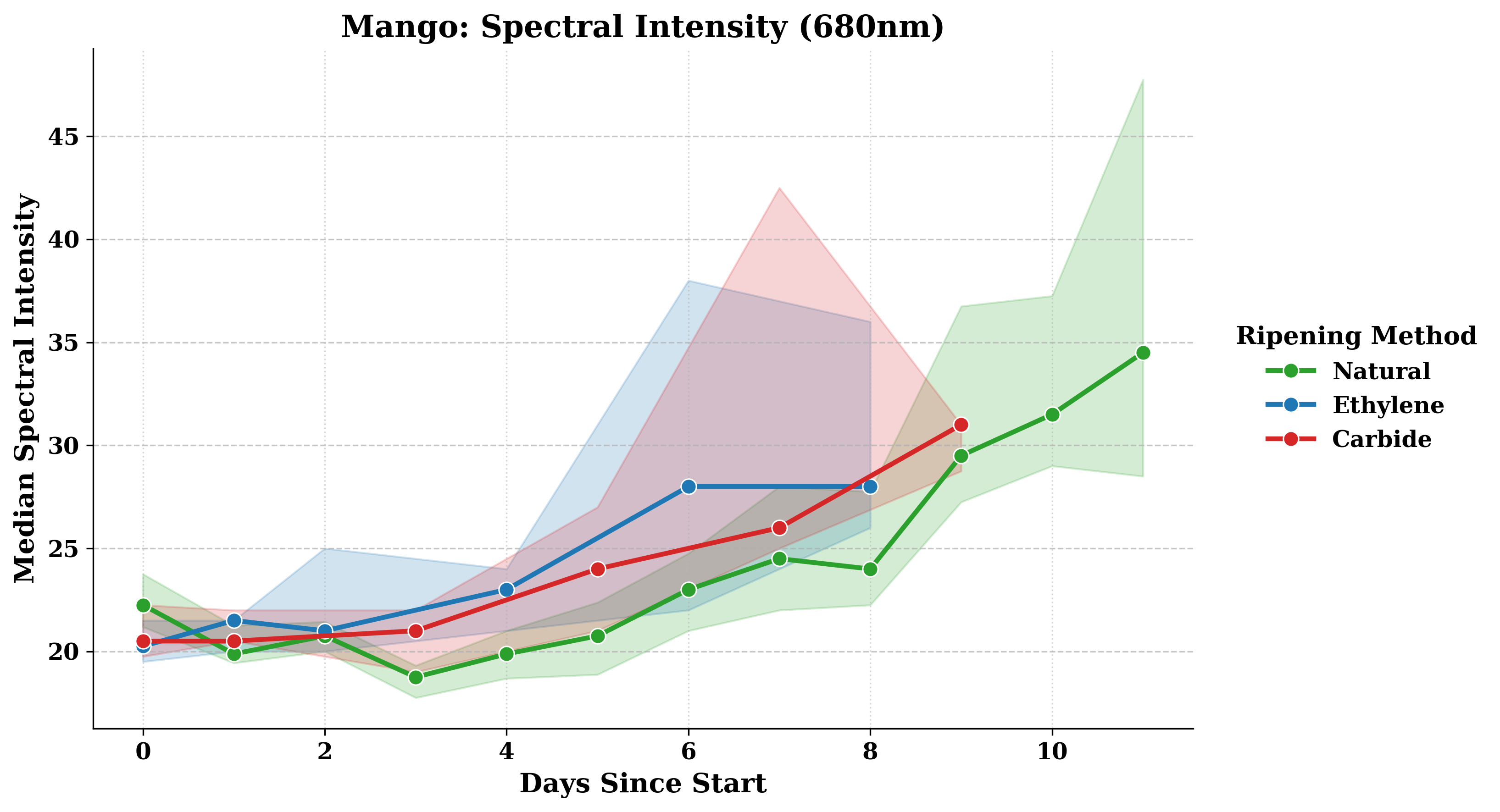}
        \caption{}
        \label{fig:mango2}
    \end{subfigure}%
    \caption{Median spectral intensity of banana samples at (a) 645nm and (b) 510nm, and mango samples at (c) 460nm and (d) 680nm, along with IQR (25th-75th percentiles) showing the overlap of spectral intensity values for all three ripening methods. Refer Table~\ref{tab:spearman_mango_fixed} and Table~\ref{tab:spearman_banana}.}
    \label{fig:mango_qual}
\end{figure}

Building on the observed spectral trends, discriminative features are extracted for differentiating between ripening methods (classification task), as well as for estimating the ripeness percentage and shelf life of a fruit sample (regression tasks). Rather than relying on raw spectral intensities alone, statistical techniques are used to identify specific wavelengths, that can be useful in both classification and regression. 
Initially, Spearman’s rank correlation coefficients between median spectral intensities and temporal progression (days since start) are computed independently across all wavelengths for each ripening method. The IQR containing the 25th-75th percentile of datapoints for each ripening method is also plotted. Wavelengths exhibiting similar Spearman's rank correlation coefficient magnitude and direction, along with overlapping IQR values; as shown in Figures~\ref{fig:banana2} and \ref{fig:mango2}, are identified as robust indicators of ripening progression, as they reflect time-dependent physiological changes, independent of the ripening method used. Conversely, wavelengths showing strong divergence in correlation behavior between the ripening methods and separated IQR ranges, are considered informative for classification. Figures~\ref{fig:banana1} and \ref{fig:mango1} are examples of such wavelengths.
The overlap between IQRs is quantified using Jaccard index, also known as IoU.

Among the 18 distinct wavelengths studied, certain wavelengths discriminate well between ripening methods, while some act as indicators of temporal ripening progression.
A wavelength is identified as an indicator of temporal progression if it demonstrates a strong monotonic correlation across natural, ethylene, and carbide ripening methods, along with a high degree of IQR overlap.
Conversely, a wavelength is identified as a discriminator of ripening methods if it exhibits low IQR overlap and diverging correlation behavior between the three ripening methods.

\begin{figure}[H]
    \centering
    \begin{subfigure}[b]{0.48\textwidth}
        \centering
        \includegraphics[width=1\textwidth, keepaspectratio]{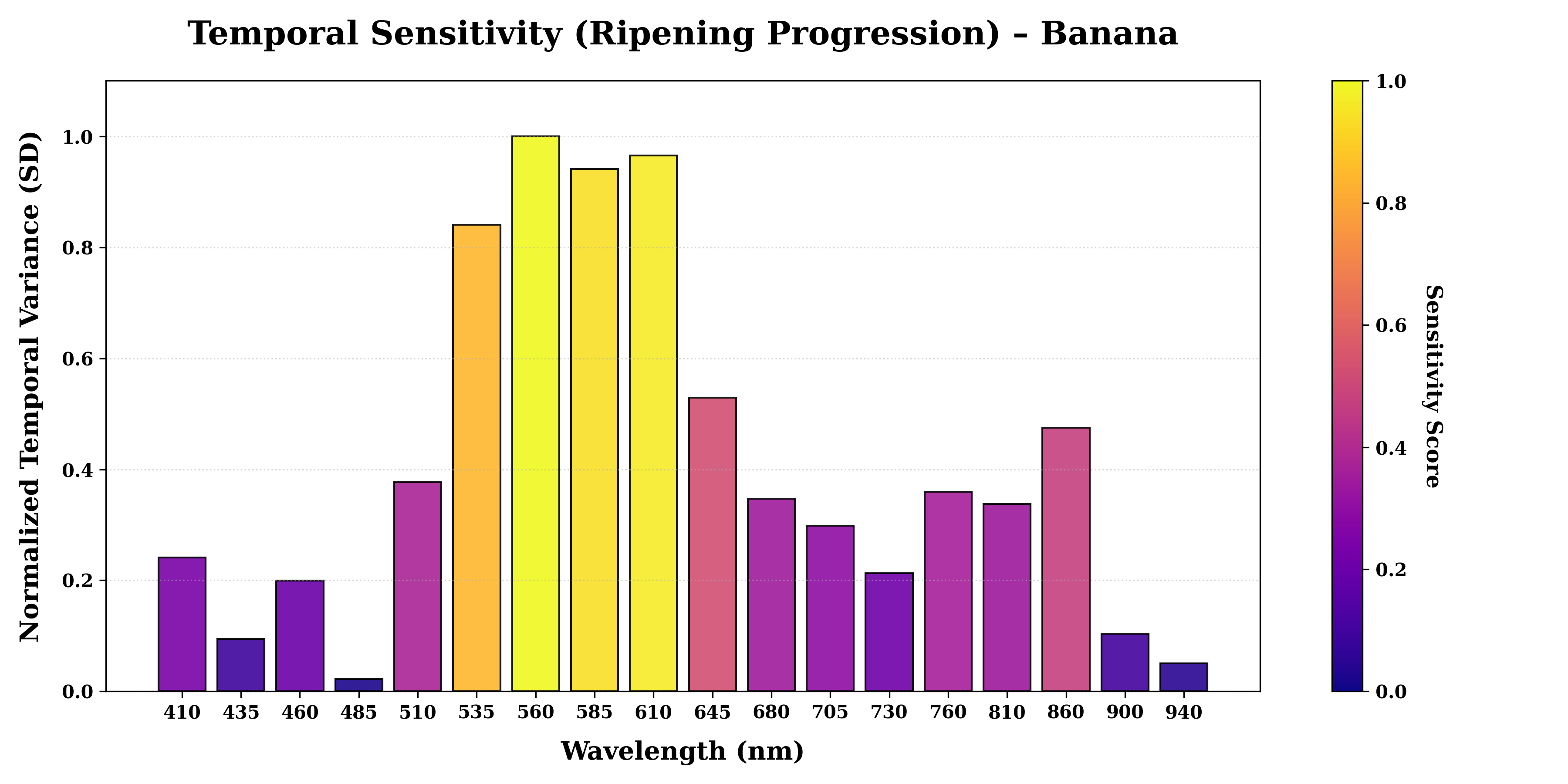} 
        \caption{}
        \label{fig:sensitivity_banana}
    \end{subfigure}%
    \vspace{0.2cm} 
    \begin{subfigure}[b]{0.48\textwidth}
        \centering
        \includegraphics[width=1\textwidth, keepaspectratio]{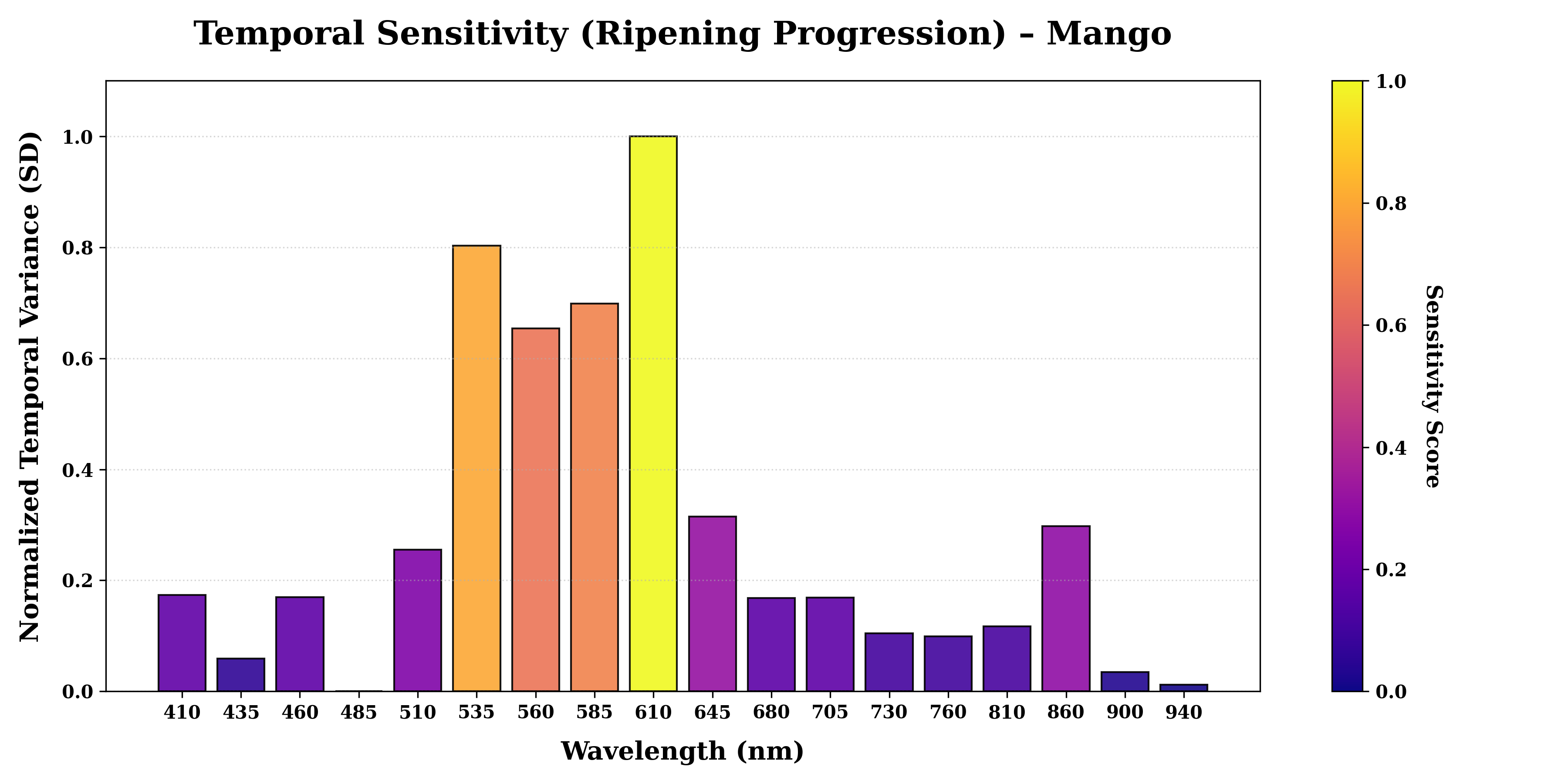}
        \caption{}
        \label{fig:sensitivity_mango}
    \end{subfigure}% 
    \vspace{0.2cm}
    \begin{subfigure}[b]{0.48\textwidth}
        \centering
        \includegraphics[width=1\textwidth, keepaspectratio]{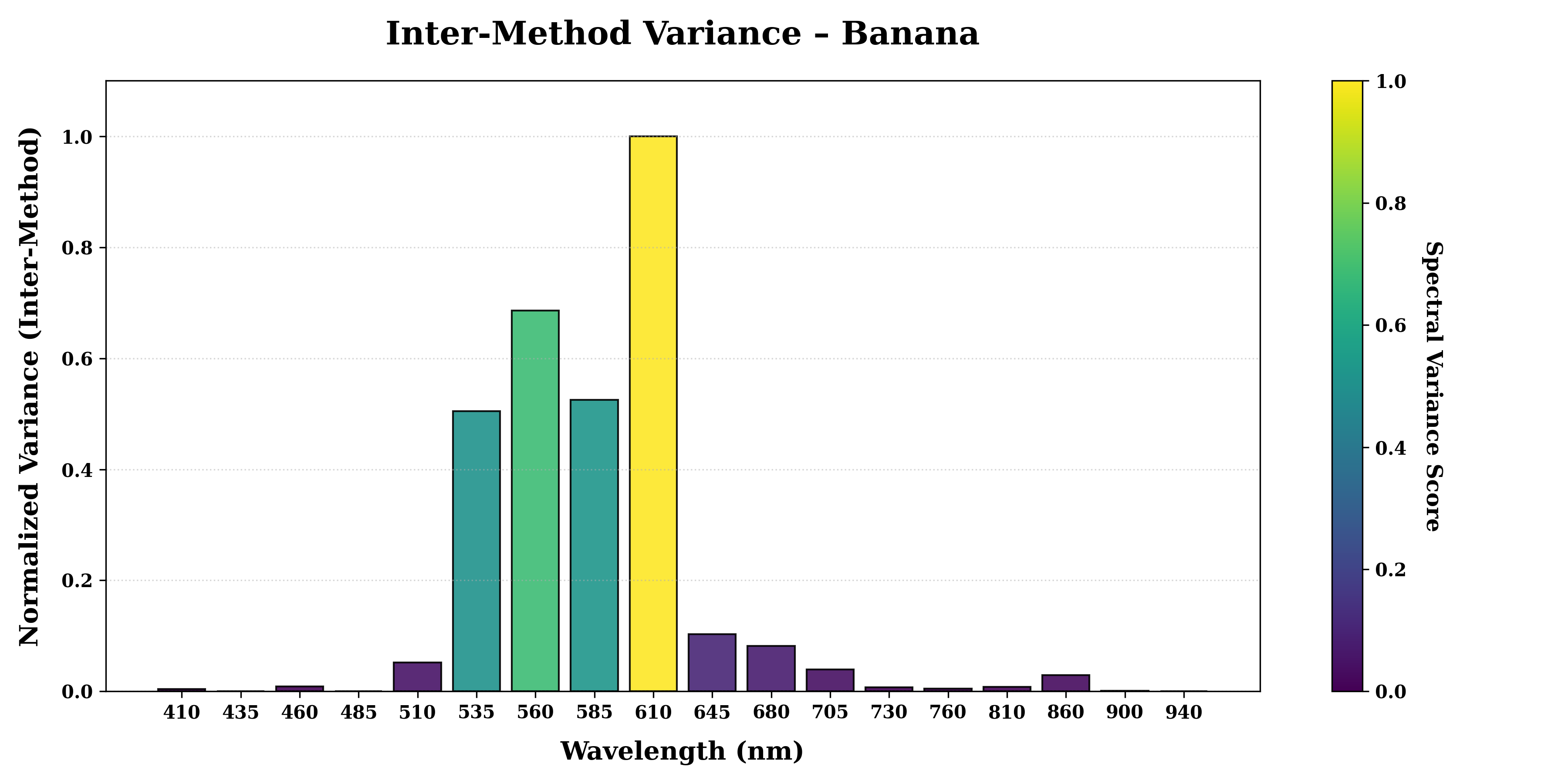}
        \caption{}
        \label{fig:variance_banana}
    \end{subfigure}%
    \vspace{0.2cm}
    \begin{subfigure}[b]{0.48\textwidth}
        \centering
        \includegraphics[width=1\textwidth, keepaspectratio]{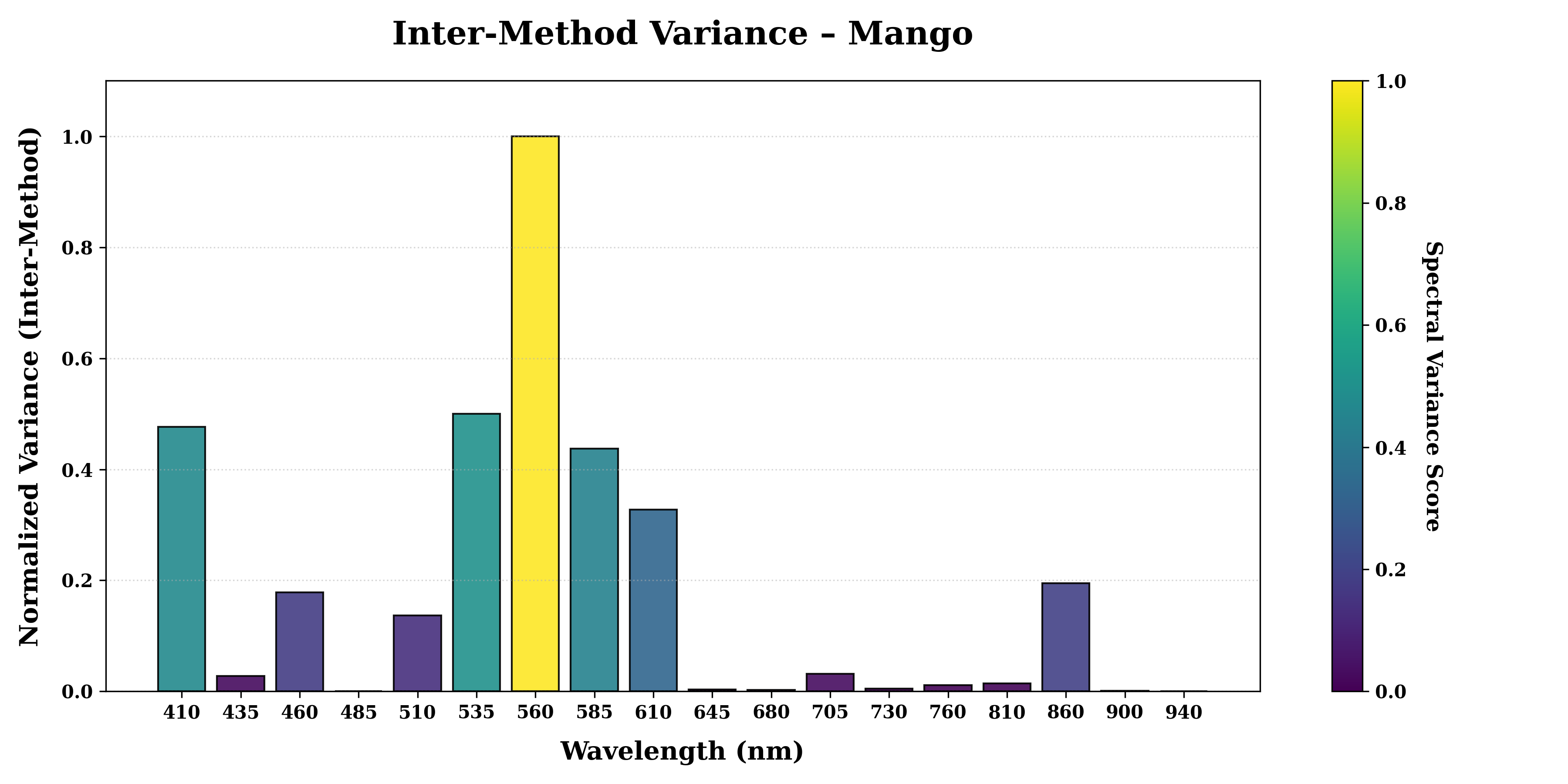}
        \caption{}
        \label{fig:variance_mango}
    \end{subfigure}%
    \caption{Temporal sensitivity and inter-method variance analysis across the 18 measured wavelengths (410–940 nm). Subplots (a) and (b) illustrate the normalized temporal variance, highlighting the spectral bands that are  most sensitive to continuous ripening progression in banana and mango samples, respectively. Subplots (c) and (d) quantify the normalized inter-method variance, illustrating the magnitude of macroscopic spectral separation between the ripening methods at each wavelength.}
    \label{fig:sensitivity}
\end{figure}

Further, in order to identify wavelengths that are more sensitive to fruit ripening over time; regardless of the method employed, standard deviation in the spectral intensity values at all 18 wavelengths is studied, and higher standard deviation is considered more informative for modelling ripening progression. As evident from Figure~\ref{fig:sensitivity}, wavelengths in visible region; specifically 535nm-610nm (green to orange/red) exhibit higher sensitivity to ripening progression of fruit.

\begin{table}[h]
    \centering
    \caption{Spearman's rank correlation coefficient values ($\rho$) of natural (N), Ethylene (E), and Carbide (C) ripening methods, along with extent of overlap of interquartile range (IQR) values in terms of IoU and the interpretation of these measures for median spectral intensity at mentioned wavelengths with respect to ``day since start'', for mango samples.}
    \label{tab:spearman_mango_fixed}
    
    \small 
    \setlength{\tabcolsep}{0.8pt} 
    \begin{tabular}{P{1.6cm} P{0.9cm} P{0.9cm} P{0.9cm} P{0.9cm}@{\hspace{0.3cm}}>{\RaggedRight\arraybackslash}p{3.3cm}}
        \toprule
        \textbf{Wavelength} & \textbf{$\rho$ (N)} & \textbf{$\rho$ (E)} & \textbf{$\rho$ (C)} & \textbf{IoU} & \multicolumn{1}{P{3.2cm}}{\textbf{Interpretation}} \\
        \midrule
        410 nm & $-$0.09 & $-$0.94 & $-$0.60 & 0.476 & Good class separability \\ \addlinespace
        460 nm & $-$0.43 & $-$0.81 & 0.77  & 0.663 & Moderate class separability due to diverging correlation behavior despite substantial IQR overlap \\ \addlinespace
        610 nm & 0.94  & 0.60  & 0.77  & 0.815 & Strong indicator of Temporal progression \\ \addlinespace
        680 nm & 0.84  & 0.93  & 0.99  & 0.917 & Strong indicator of Temporal progression \\
        \bottomrule
    \end{tabular}
\end{table}
\begin{table}[H]
    \centering
    \caption{Spearman's rank correlation coefficient values ($\rho$) of natural (N), Ethylene (E), and Carbide (C) ripening methods, along with extent of overlap of interquartile range (IQR) values in terms of IoU and the interpretation of these measures for median spectral intensity at mentioned wavelengths with respect to ``day since start'', for banana samples.}
    \label{tab:spearman_banana}
    
    \small 
    \setlength{\tabcolsep}{0.8pt} 
    \begin{tabular}{P{1.6cm} P{0.9cm} P{0.9cm} P{0.9cm} P{0.9cm}@{\hspace{0.3cm}}>{\RaggedRight\arraybackslash}p{3.3cm}}
        \toprule
        \textbf{Wavelength} & \textbf{$\rho$ (N)} & \textbf{$\rho$ (E)} & \textbf{$\rho$ (C)} & \textbf{IoU} & \multicolumn{1}{P{3.2cm}}{\textbf{Interpretation}} \\
        \midrule
        460 nm & $-$0.99 & $-$0.80 & $-$0.89 & 0.800 & Strong indicator of temporal progression \\ \addlinespace
        510 nm & $-$0.88 & $-$0.90 & $-$1.00 & 0.569 & Moderate temporal progression indicator due to strong monotonic correlation but limited IQR overlap \\ \addlinespace
        645 nm & 0.32  & $-$0.10 & $-$0.60 & 0.821 & Limited class separability due to high IQR overlap and weak/diverging temporal correlations \\ \addlinespace
        730 nm & $-$0.38 & $-$1.00 & $-$0.94 & 0.989 & Strong Indicator of temporal progression \\
        \bottomrule
    \end{tabular}
\end{table}

\begin{figure}[H]
    \centering
    % --- Image 1 ---
    
    \begin{subfigure}[b]{1.0\linewidth}
        \centering
        \includegraphics[width=\linewidth, height=4.5cm, keepaspectratio]{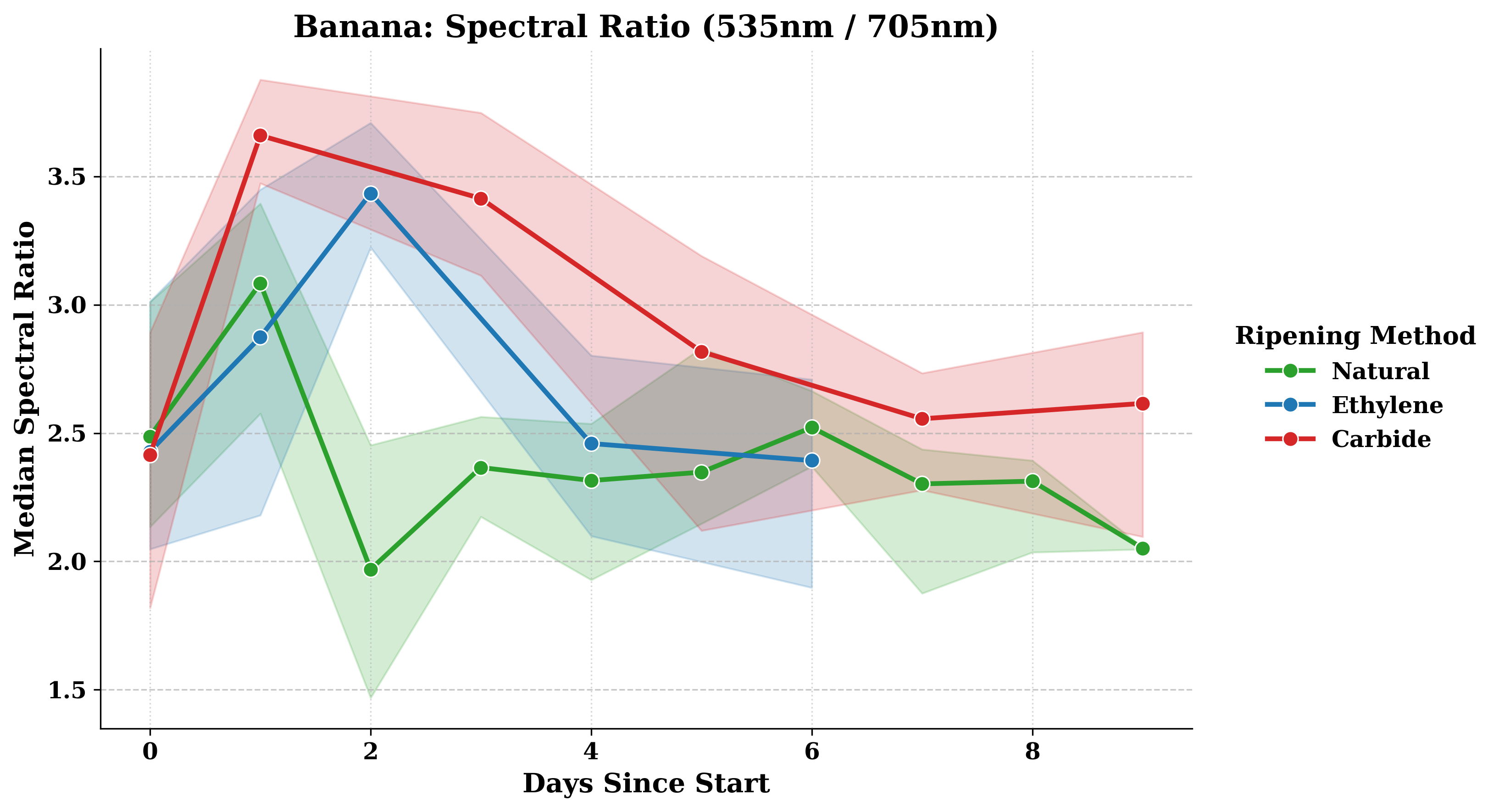} 
        \caption{}
        \label{fig:banana1_ratio}
    \end{subfigure}%
    \vspace{0.15cm}
    % --- Image 2 ---
    \begin{subfigure}[b]{1.0\linewidth}
        \centering
        \includegraphics[width=\linewidth, height=4.5cm, keepaspectratio]{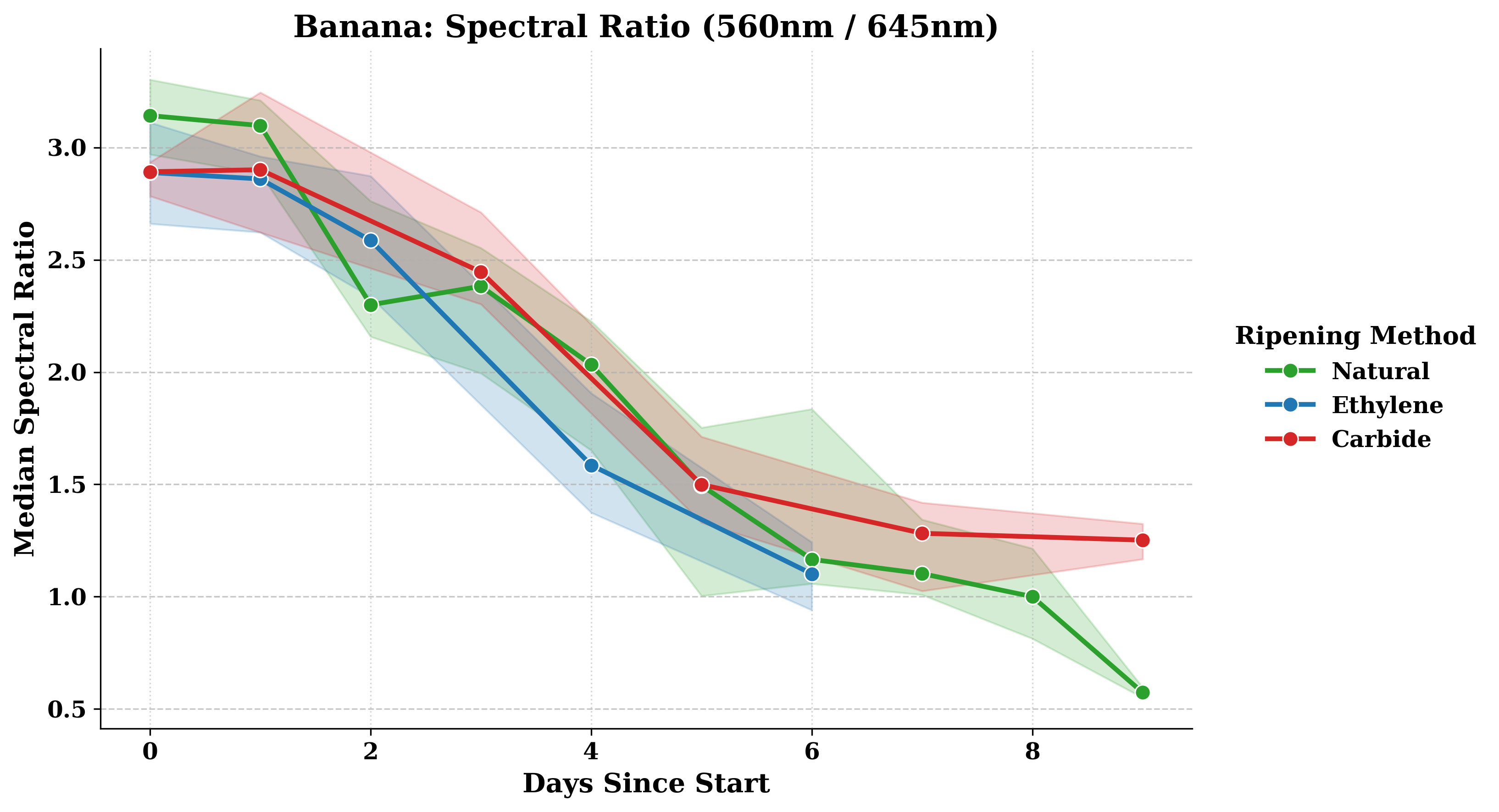}
        \caption{}
        \label{fig:banana2_ratio}
    \end{subfigure}%
    \vspace{0.15cm}
    % --- Image 3 ---
    \begin{subfigure}[b]{1.0\linewidth}
        \centering
        \includegraphics[width=\linewidth, height=4.5cm, keepaspectratio]{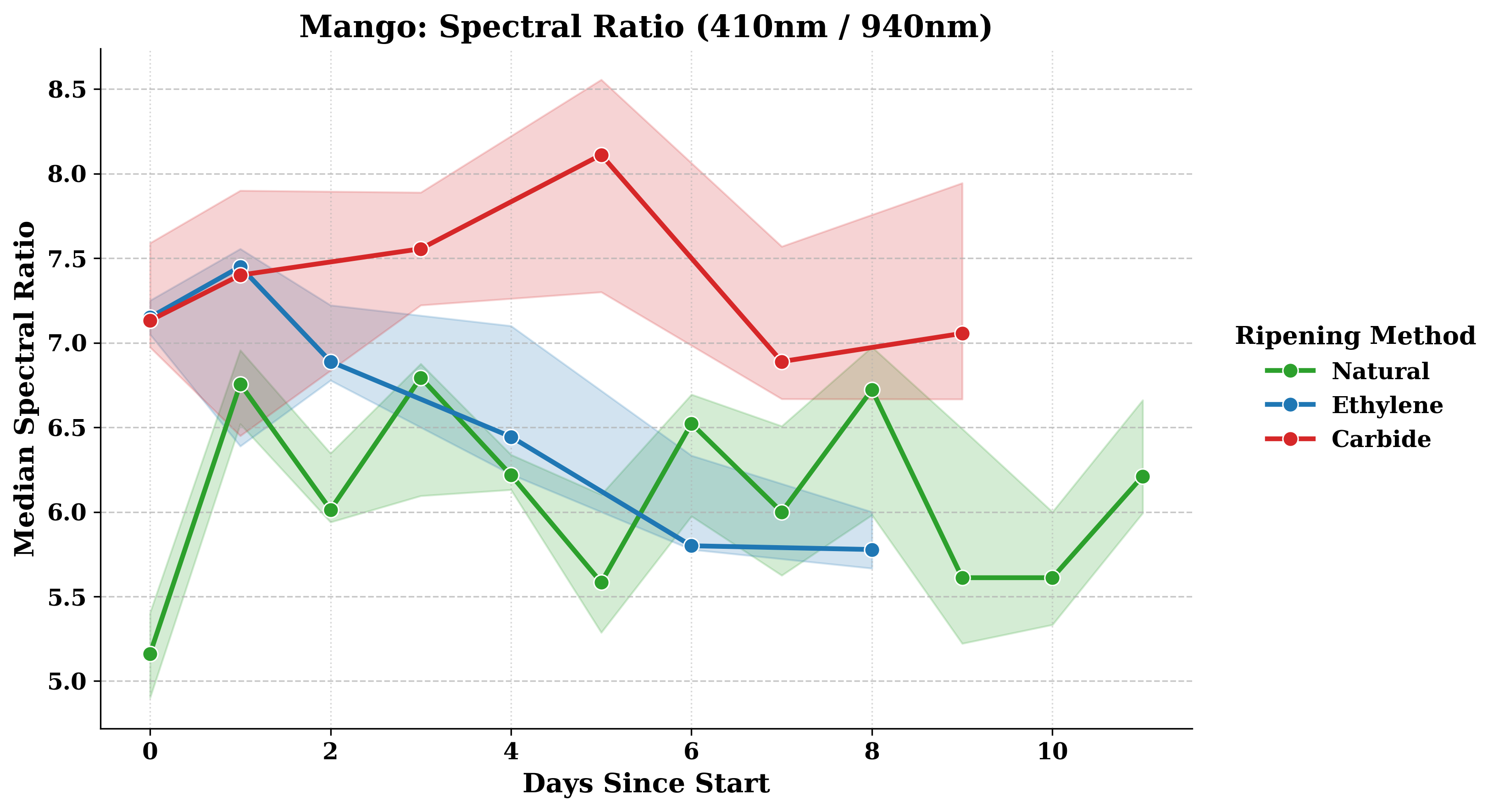}
        \caption{}
        \label{fig:mango1_ratio}
    \end{subfigure}%
    \vspace{0.15cm}
    % --- Image 4 ---
    \begin{subfigure}[b]{1.0\linewidth}
        \centering
        \includegraphics[width=\linewidth, height=4.5cm, keepaspectratio]{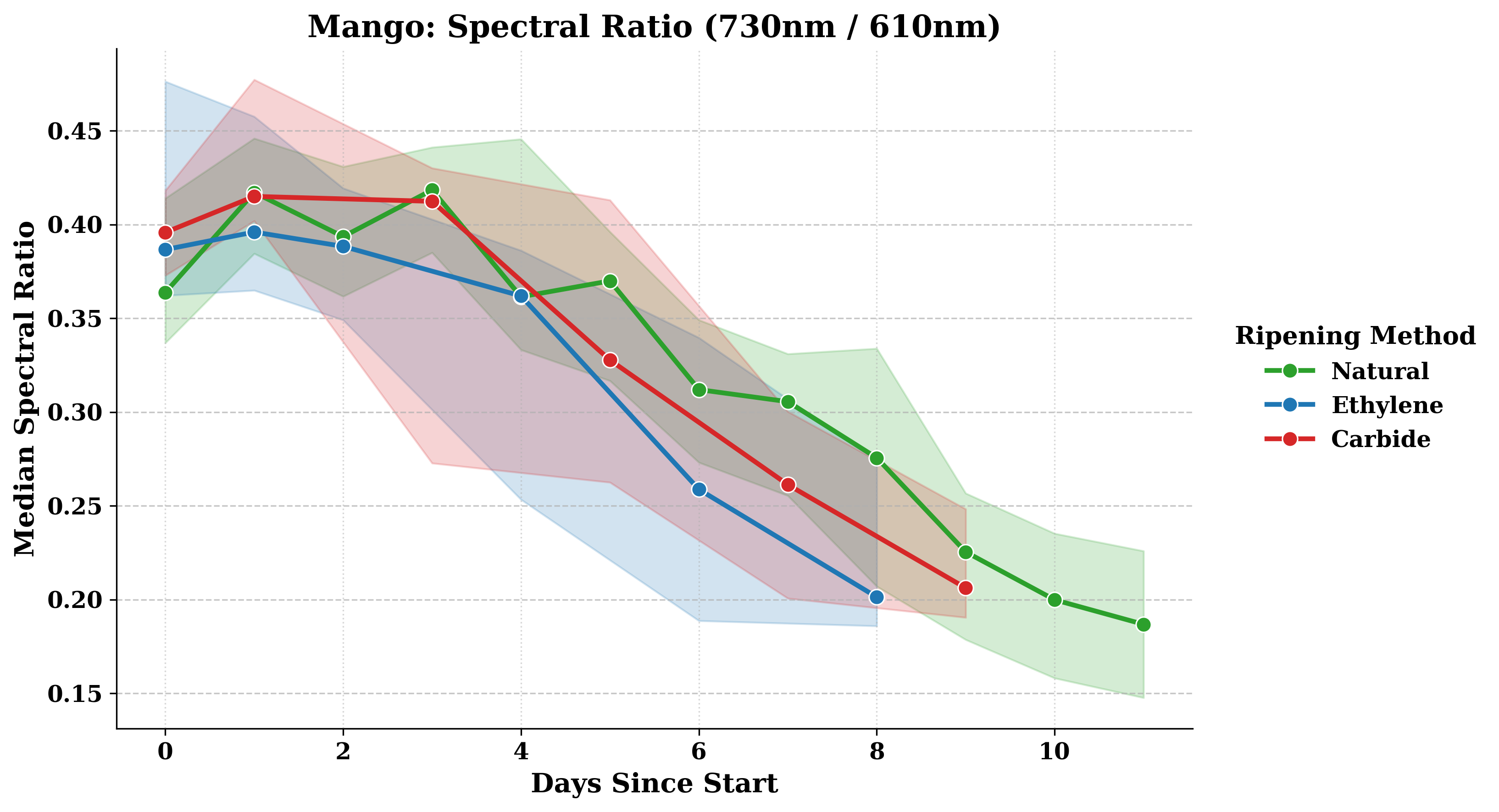}
        \caption{}
        \label{fig:mango2_ratio}
    \end{subfigure}%
    \caption{Ratios of median spectral intensity of banana samples at (a) 535 nm / 705 nm and (b) 560 nm / 645 nm, and mango samples at (c) 410 nm / 940 nm and (d) 730 nm / 610 nm , along with IQR (25th-75th percentiles) showing the overlap of spectral intensity values, for all three ripening methods. Refer Table~\ref{tab:spearman_ratios_mango} and Table~\ref{tab:spearman_ratios_banana}.} 
    \label{fig:ratio_progression}
\end{figure}

Similarly, the inter-method variance of spectral intensity values at all 18 wavelengths is studied. The wavelengths exhibiting higher variance (560 nm for mango, and 610 nm for banana), differ the most in intensity values across the three methods. These differences are consistent with the pigment-related spectral changes expected during ripening, although the present measurements do not directly quantify pigment concentrations (Figure~\ref{fig:wavelength_progression}).

In order to amplify relative changes and minimize variations caused by sensor drift or illumination differences, spectral ratios are explored by evaluating all pairwise wavelength ratios. A total of 306 such ratios (since 18 wavelengths) are explored and their relevance checked based on spearman's rank correlation coefficient values and IQR ranges. Figures~\ref{fig:banana2_ratio} and \ref{fig:mango2_ratio} show ratios with similar Spearman's rank correlation coefficients and overlapping IQR values for banana and mango respectively, making them a good indicator of ripening progression and hence useful for regression.
Similarly, Figures~\ref{fig:banana1_ratio} and \ref{fig:mango1_ratio} depict ratios with diverging correlation values and separated IQR for the three methods for banana and mango samples, respectively. These wavelengths discriminate well between the three ripening methods used in the study. The ratios which act as either good indicators of ripening progression or discriminate well between ripening methods are presented in Table~\ref{tab:spearman_ratios_mango} for mango samples and Table~\ref{tab:spearman_ratios_banana} for banana samples, along with their Spearman’s rank correlation coefficient values and IQR characteristics.
\begin{table}[H]
    \centering
    \caption{Spearman's rank correlation coefficient values ($\rho$) of natural (N), Ethylene (E), and Carbide (C) ripening methods, along with extent of overlap of interquartile range (IQR) values in terms of IoU and the interpretation of these measures for given ratios of median spectral intensity with respect to ``day since start'', for mango samples.}
    \label{tab:spearman_ratios_mango}
    
    \small 
    \setlength{\tabcolsep}{1.5pt} 
    \begin{tabular}{>{\RaggedRight\arraybackslash}p{1.3cm} P{0.9cm} P{0.9cm} P{0.9cm} P{0.9cm}@{\hspace{0.3cm}}>{\RaggedRight\arraybackslash}p{3.3cm}}
        \toprule
        \multicolumn{1}{P{1.2cm}}{\textbf{Ratios}} & \textbf{$\rho$ (N)} & \textbf{$\rho$ (E)} & \textbf{$\rho$ (C)} & \textbf{IoU} & \multicolumn{1}{P{3.3cm}}{\textbf{Interpretation}} \\
        \midrule
        410 nm / 940 nm & $-$0.11 & $-$0.94 & $-$0.37 & 0.373  & Good class separability \\ \addlinespace
        730 nm / 610 nm & $-$0.90 & $-$0.83 & $-$0.83 & 0.932 & Strong indicator of temporal progression \\ \addlinespace
        610 nm / 705 nm & 0.90  & 0.94  & 1.00  & 0.682 & Moderate temporal progression indicator due to strong correlation but limited IQR overlap \\
        \bottomrule
    \end{tabular}
\end{table}

\begin{table}[H]
    \centering
    \caption{Spearman's rank correlation coefficient values ($\rho$) of natural (N), Ethylene (E), and Carbide (C) ripening methods, along with extent of overlap of interquartile range (IQR) values in terms of IoU and the interpretation of these measures for given ratios of median spectral intensity with respect to ``day since start'', for banana samples.}
    \label{tab:spearman_ratios_banana}
    
    \small 
    \setlength{\tabcolsep}{1.5pt} 
    \begin{tabular}{>{\RaggedRight\arraybackslash}p{1.3cm} P{0.9cm} P{0.9cm} P{0.9cm} P{0.9cm}@{\hspace{0.3cm}}>{\RaggedRight\arraybackslash}p{3.3cm}}
        \toprule
        \multicolumn{1}{P{1.2cm}}{\textbf{Ratios}} & \textbf{$\rho$ (N)} & \textbf{$\rho$ (E)} & \textbf{$\rho$ (C)} & \textbf{IoU} & \multicolumn{1}{P{3.3cm}}{\textbf{Interpretation}} \\
        \midrule
        585 nm / 645 nm & $-$0.98 & $-$1.00 & $-$1.00 & 0.923 & Strong indicator of temporal progression \\ \addlinespace
        560 nm / 645 nm & $-$0.99 & $-$1.00 & $-$0.94 & 0.912 & Strong indicator of temporal progression \\ \addlinespace
        535 nm / 705 nm & $-$0.77 & $-$0.60 & $-$0.37 & 0.744 & Moderate class separability \\
        \bottomrule
    \end{tabular}
\end{table}
\begin{figure}[H]
    \centering
    \begin{subfigure}[b]{0.48\textwidth}
        \centering
        \includegraphics[height=3.9cm, keepaspectratio]{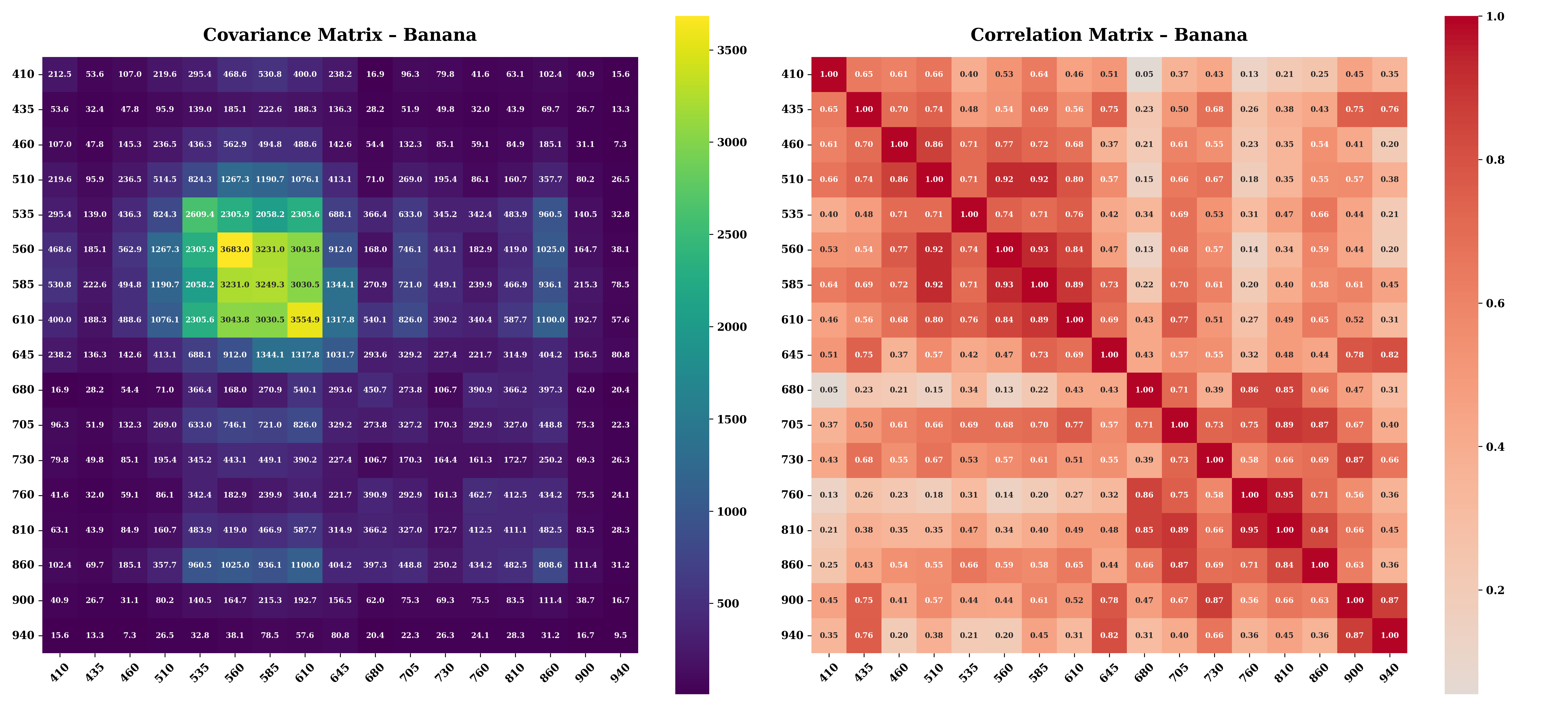} 
        \caption{Covariance matrix (left) and correlation matrix (right) for banana}
        \label{fig:covariance_banana}
    \end{subfigure}% <--- Percent signs prevent unwanted line breaks
    \vspace{0.2cm} 
    \begin{subfigure}[b]{0.48\textwidth}
        \centering
        \includegraphics[height=3.9cm, keepaspectratio]{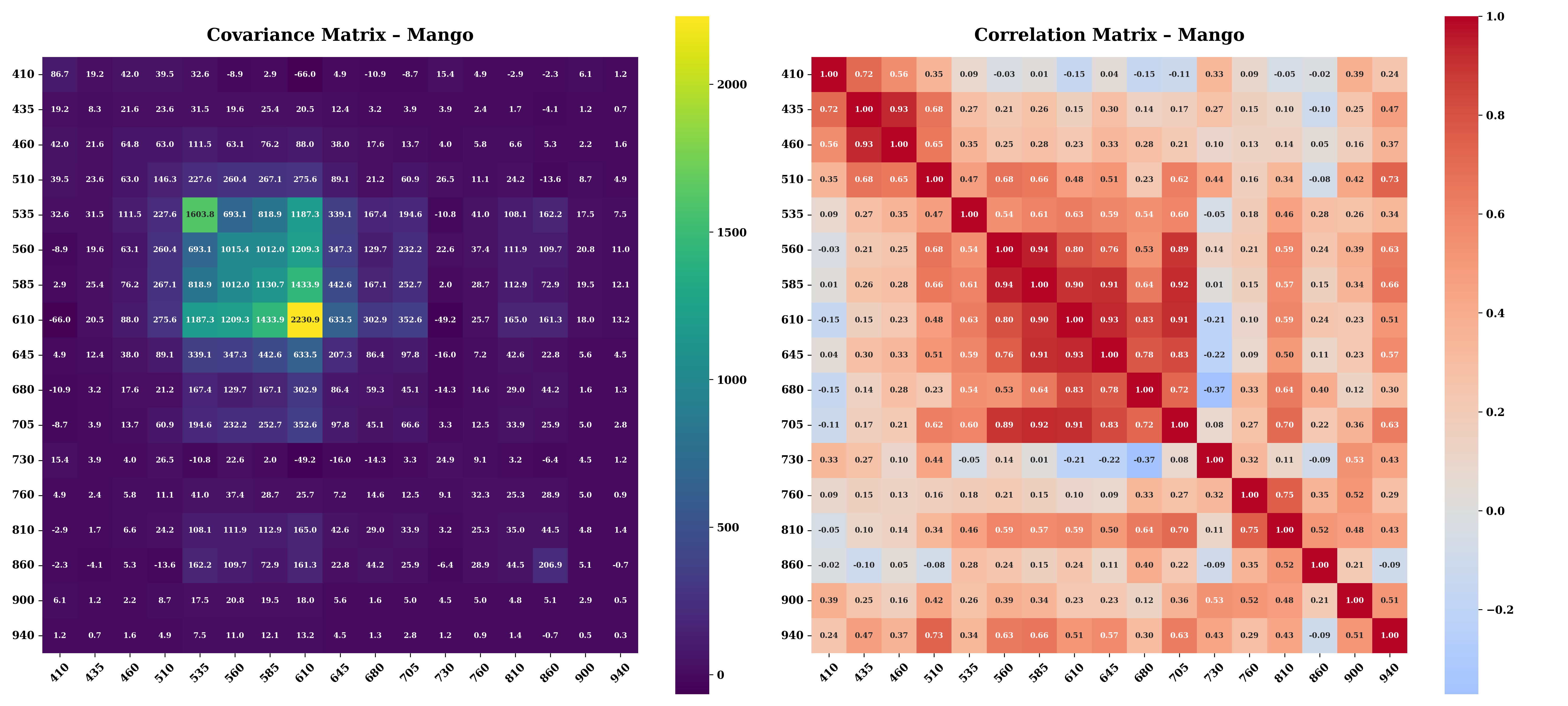}
        \caption{Covariance matrix (left) and correlation matrix (right) for mango}
        \label{fig:covariance_mango}
    \end{subfigure}%
    \caption{Covariance (left) and correlation (right) matrices of raw spectral intensities across the 18 measured wavelengths for (a) banana and (b) mango samples. The heatmaps illustrate a high degree of collinearity among adjacent spectral bands, particularly within the visible and NIR regions, justifying the need for dimensionality reduction.}
    \label{fig:pca}
\end{figure}
\begin{figure}[H]
    \centering
    \begin{subfigure}[b]{0.48\textwidth}
        \centering
        \includegraphics[height=4.5cm, keepaspectratio]{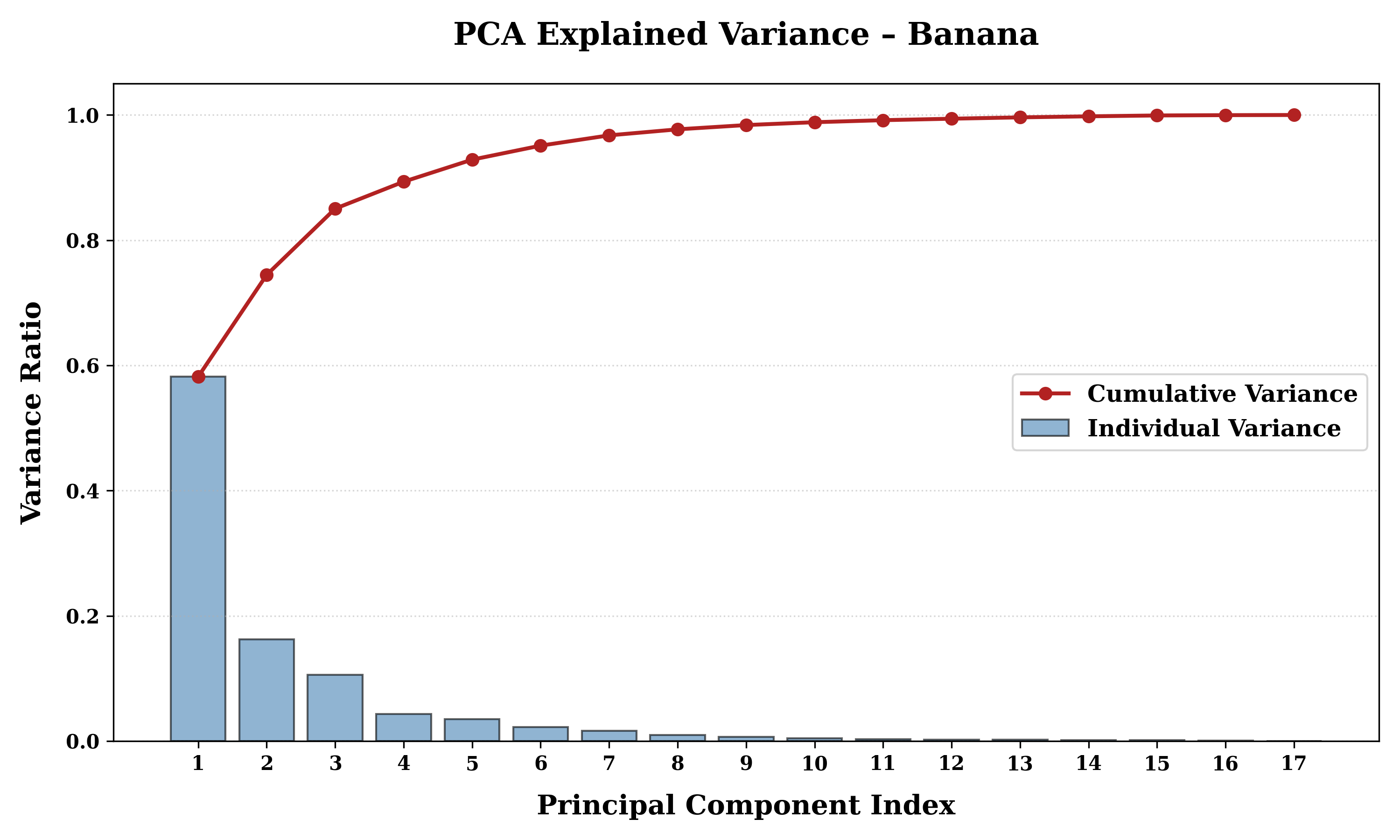} 
        \caption{}
        \label{fig:explained_variance_banana}
    \end{subfigure}% <--- Percent signs prevent unwanted line breaks
    \vspace{0.1cm} 
    \begin{subfigure}[b]{0.48\textwidth}
        \centering
        \includegraphics[height=4.5cm, keepaspectratio]{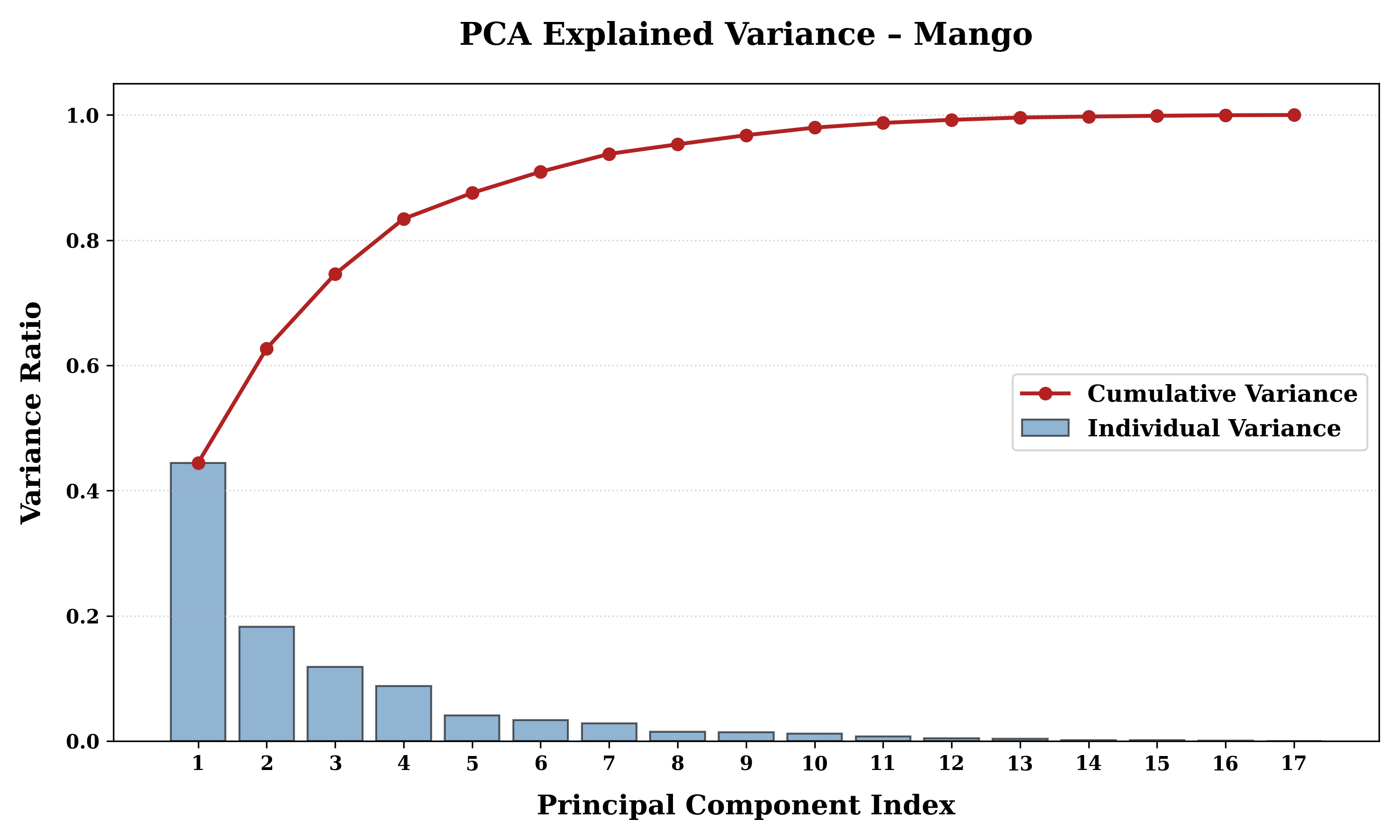}
        \caption{}
        \label{fig:explained_variance_mango}
    \end{subfigure}%
    \caption{Individual explained variance (bars) and cumulative variance (line) ratios of the principal components derived from the spectral data of (a) banana and (b) mango. The scree plots clearly show that the first few principal components capture majority of the spectral variance, enabling dimensionality reduction without significant information loss.}
    \label{fig:explained_variance}
\end{figure}

\begin{figure}[H]
    \centering
    \begin{subfigure}[b]{0.48\textwidth}
        \centering
        \includegraphics[width=1\textwidth, keepaspectratio]{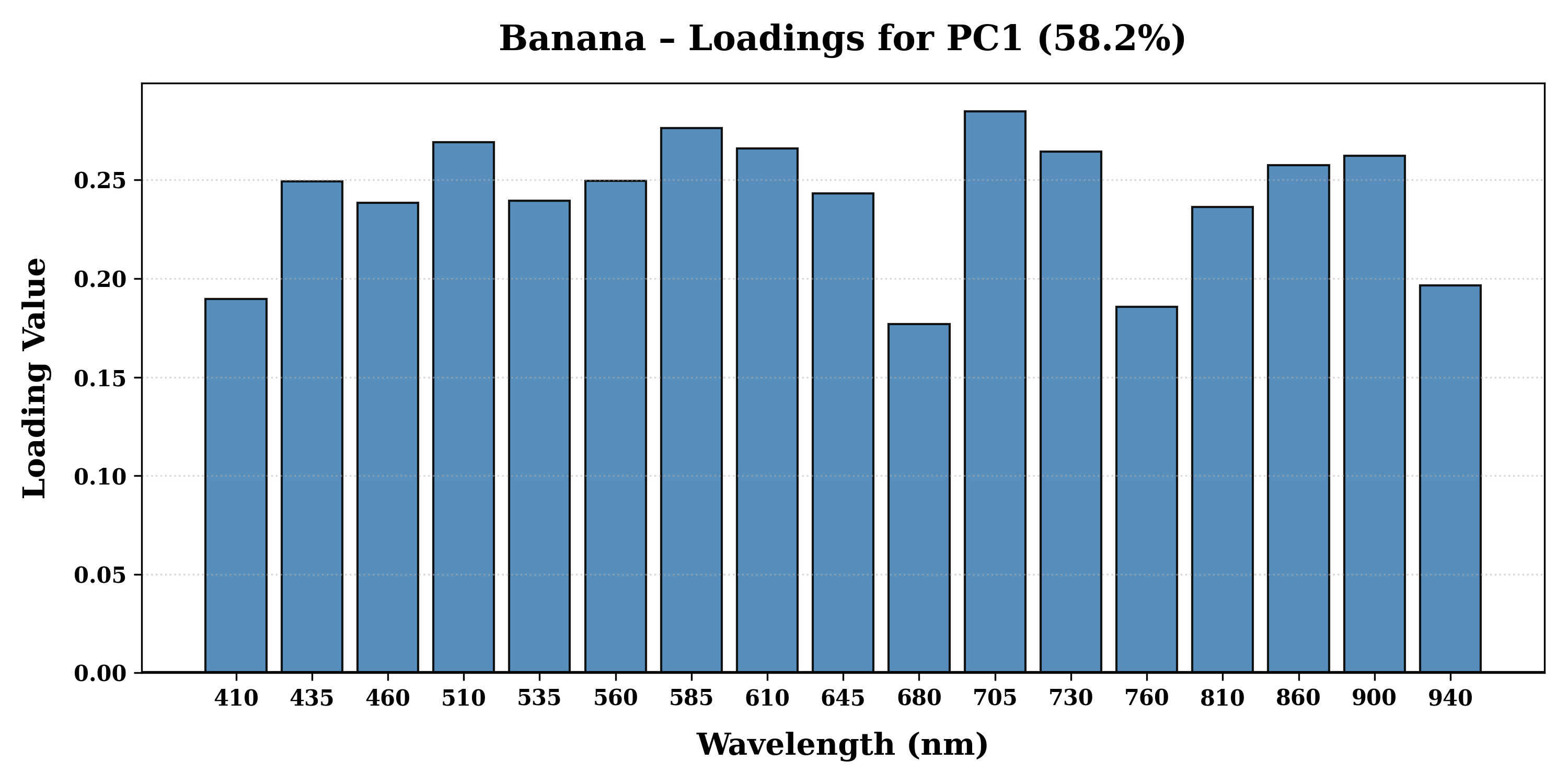} 
        \caption{}
        \label{fig:loadings_pc1_banana}
    \end{subfigure}% 
    \vspace{0.2cm} 
    \begin{subfigure}[b]{0.48\textwidth}
        \centering
        \includegraphics[width=1\textwidth, keepaspectratio]{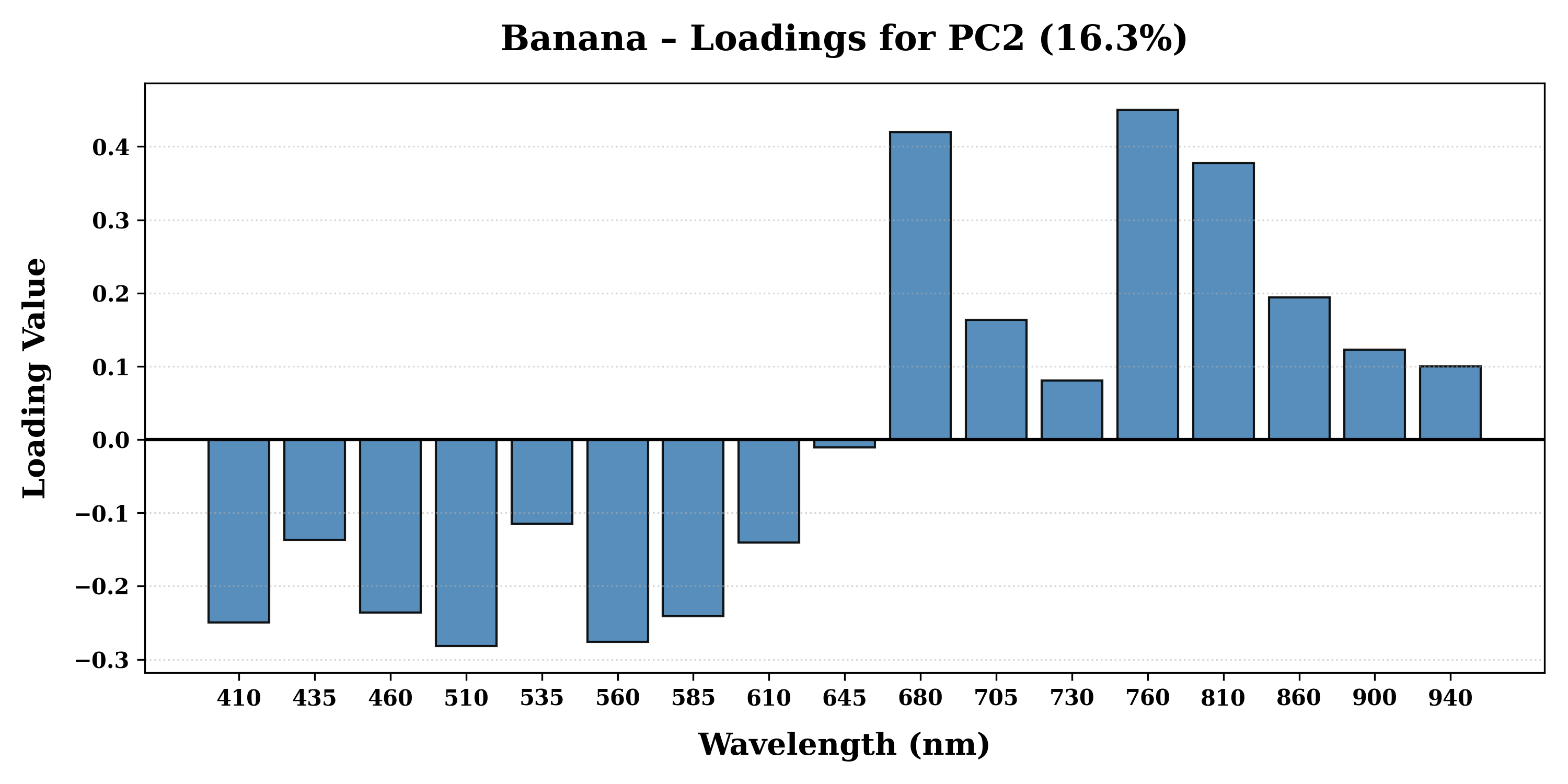}
        \caption{}
        \label{fig:loadings_pc2_banana}
    \end{subfigure}%
    \vspace{0.2cm}
    \begin{subfigure}[b]{0.48\textwidth}
        \centering
        \includegraphics[width=1\textwidth, keepaspectratio]{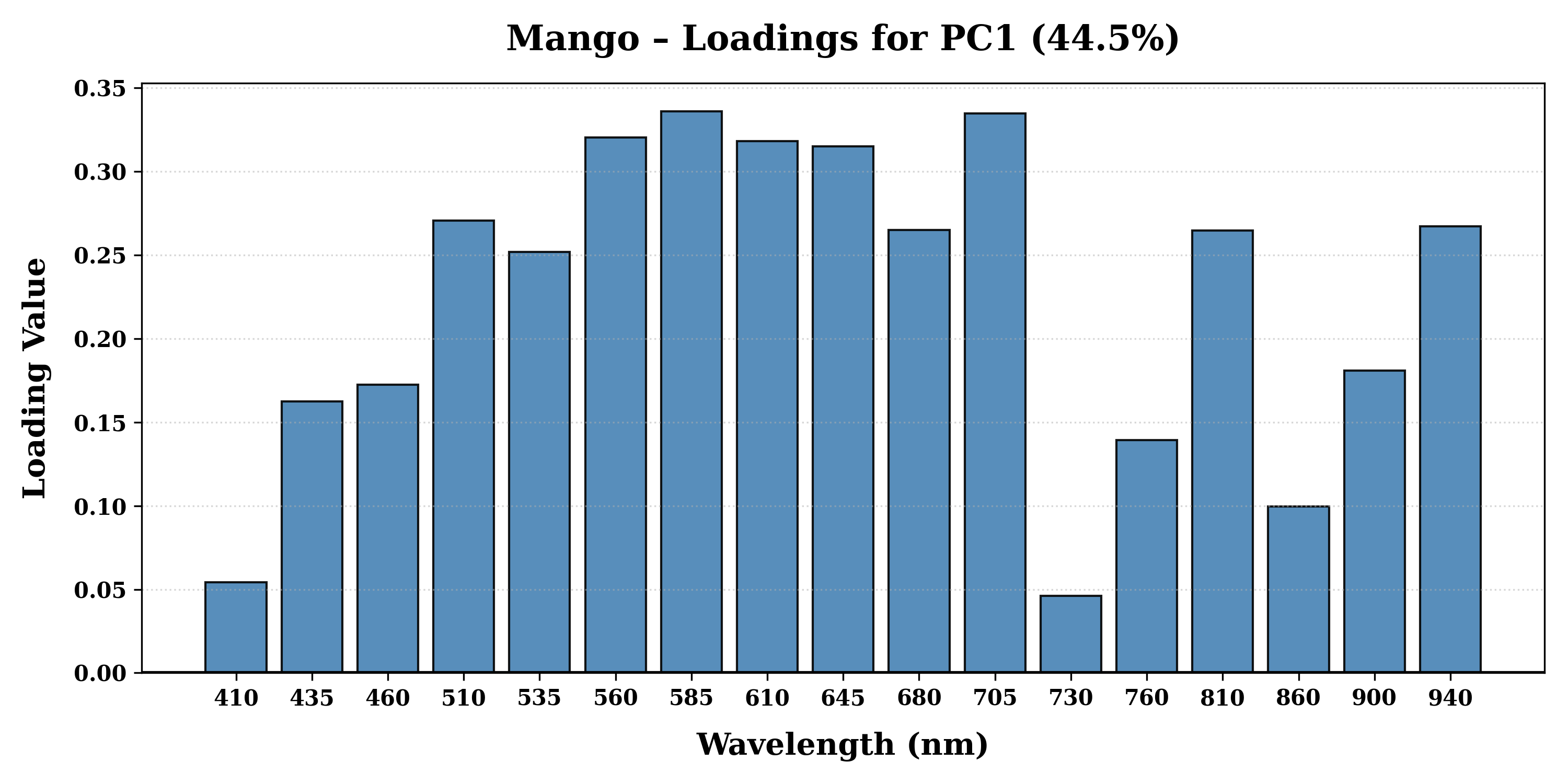}
        \caption{}
        \label{fig:loadings_pc1_mango}
    \end{subfigure}%
    \vspace{0.2cm}
    \begin{subfigure}[b]{0.48\textwidth}
        \centering
        \includegraphics[width=1\textwidth, keepaspectratio]{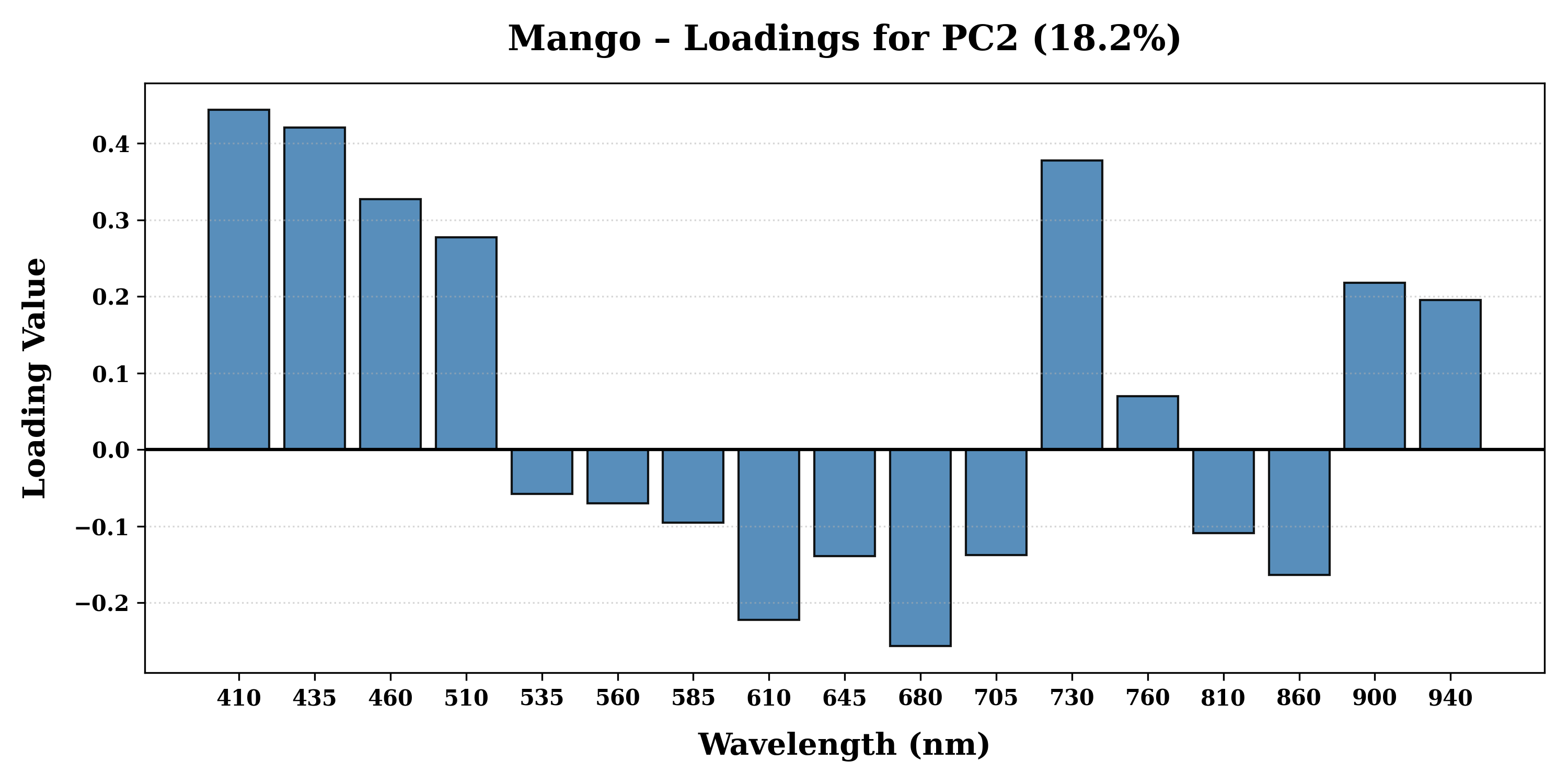}
        \caption{}
        \label{fig:loadings_pc2_mango}
    \end{subfigure}%
    \caption{PCA spectral loading values across 18 distinct wavelengths (410–940 nm). Panels (a) and (c) present the first principal component (PC1) for banana and mango samples, while panels (b) and (d) display their corresponding second principal component (PC2).}
    \label{fig:loading}
\end{figure}
\subsection{\textbf{Dimensionality Reduction}}
While individual wavelengths and ratios provide valuable information, it is important to study how the spectral intensities at all 18 wavelengths are correlated, in order to reduce the dimensionality of the data and make the learning algorithm less prone to overfitting.
Hence, covariance and correlation matrices are computed for spectral intensity values at all wavelengths. 
As seen in Figure~\ref{fig:pca}, the covariance matrices show extensive positive correlation among adjacent wavelengths, particularly within the visible and near-infrared regions. This confirms redundancy in the spectral data and hence the need of dimensionality reduction, for which PCA is employed. Eigenvalue decomposition of the covariance matrices shows a steep decay, with the first few principal components capturing the majority of spectral variance. Explained variance ratio and cumulative variance plots (Figure~\ref{fig:explained_variance}) consistently indicate that the first 5-7 principal components account for more than 90\% of the total spectral variance. Analysis of the Loading values (Figure~\ref{fig:loading}) reveals that the first two principal components are dominated by contributions from visible and NIR wavelengths, aligning with chlorophyll breakdown and carotenoid formation in the visible region and changes in spectral intensity patterns in the NIR band, which is associated with softening of the cell walls and moisture loss \citep{kapoor2022}.

\begin{figure}[H]    
    \centering
    \begin{subfigure}[b]{0.48\textwidth}
        \centering
        \includegraphics[width=1\textwidth, keepaspectratio]{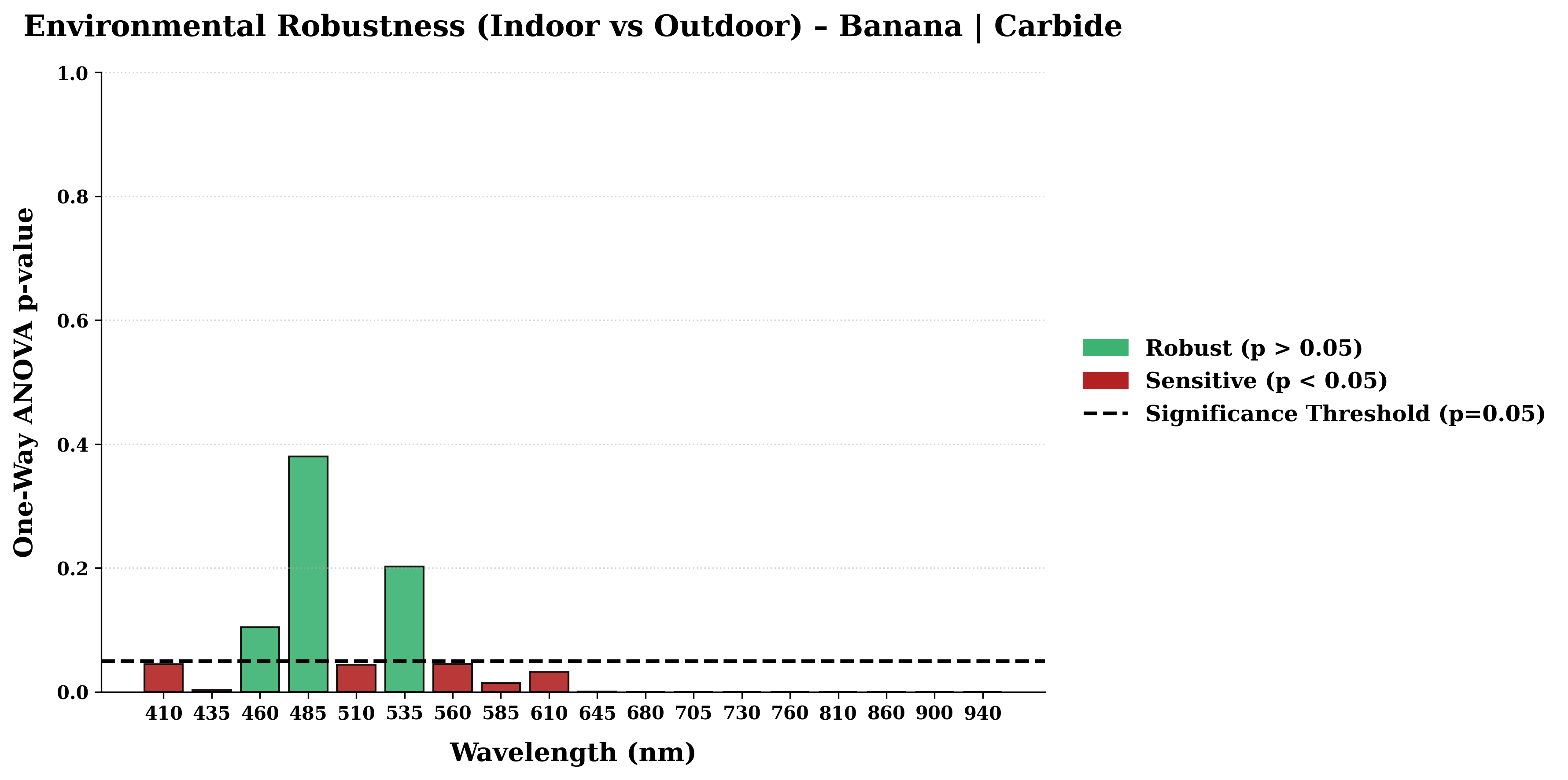} 
        \caption{}
        \label{fig:robustness_banana_carbide}
    \end{subfigure}%
    \vspace{0.2cm} 
    \begin{subfigure}[b]{0.48\textwidth}
        \centering
        \includegraphics[width=1\textwidth, keepaspectratio]{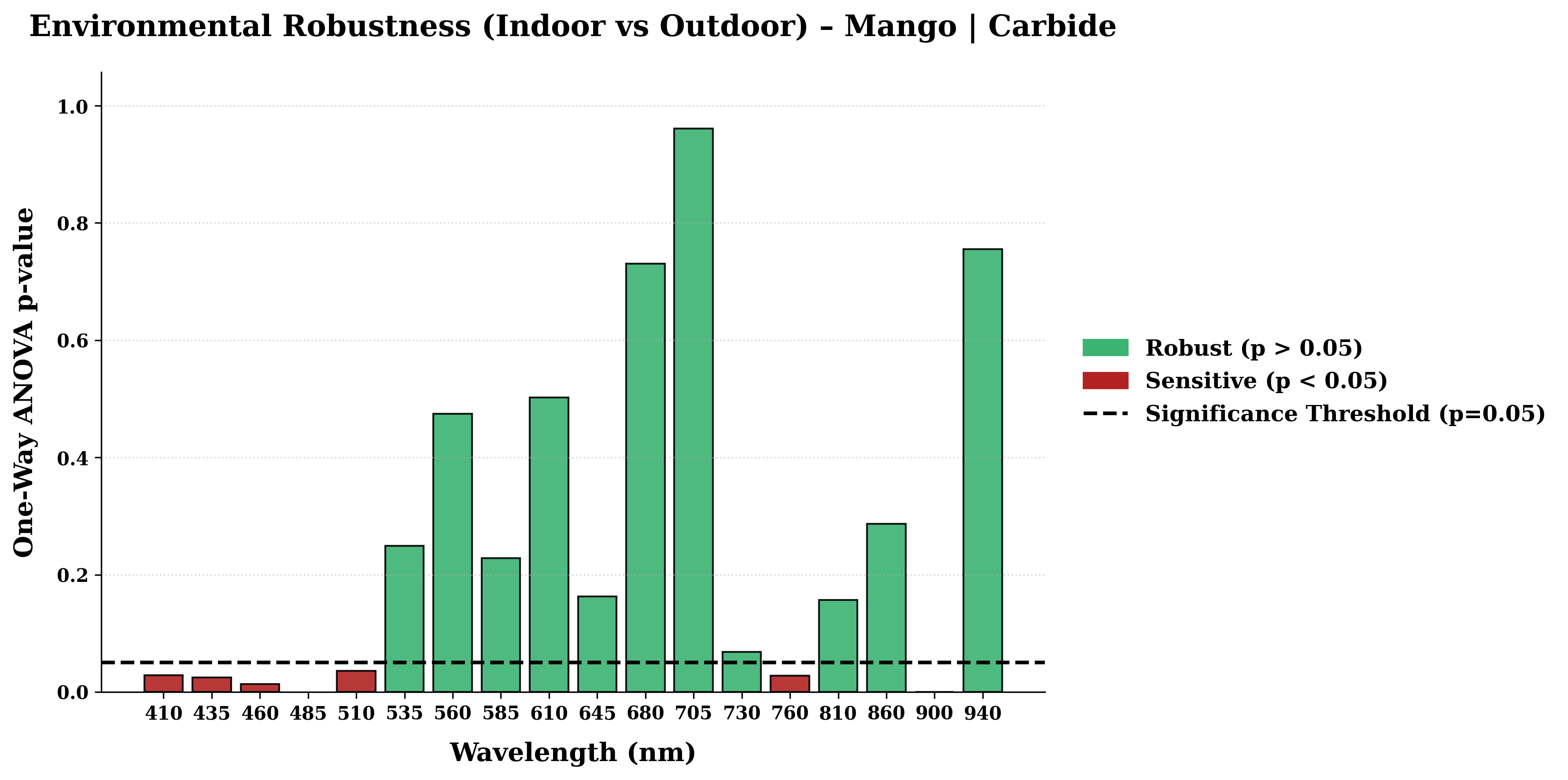}
        \caption{}
        \label{fig:robustness_mango_carbide}
    \end{subfigure}%    
    \caption{Analysis of environmental robustness of spectral intensity at all 18 wavelengths using one-way ANOVA. The bar charts display the $p$-values obtained by comparing spectral intensity variations between indoor and outdoor conditions for (a) calcium carbide-ripened bananas and (b) calcium carbide-ripened mangoes. The dashed significance threshold ($p = 0.05$) separates environmentally robust wavelengths (green bars, $p > 0.05$) from those more sensitive to temperature and humidity. (red bars, $p < 0.05$).}
    \label{fig:environmental}
\end{figure}

\subsection{\textbf{Environmental Calibration}}
To take the impact of environmental conditions into account, one-way ANOVA is performed for each wavelength by comparing the spectral intensity values in indoor and outdoor conditions, within the same fruit and ripening method combinations. Wavelengths with a p-value below 0.05 are considered to exhibit significant differences in spectral intensity between the indoor and outdoor conditions. As observed in Figure~\ref{fig:environmental}, the majority of the wavelengths for CaC\textsubscript{2}-ripened bananas exhibit significant sensitivity to environmental conditions. Conversely, the 535 nm to 940 nm range for CaC\textsubscript{2}-ripened mangoes remains largely robust against these variations. This suggests that the spectral profiles of chemically ripened bananas are inherently more susceptible to temperature and humidity fluctuations than those of mangoes.

\subsection{\textbf{Training the Learning Algorithm}}
\begin{figure}[H]
    \centering
    \begin{subfigure}[b]{0.48\textwidth}
        \centering
        \includegraphics[width=1\textwidth, keepaspectratio]{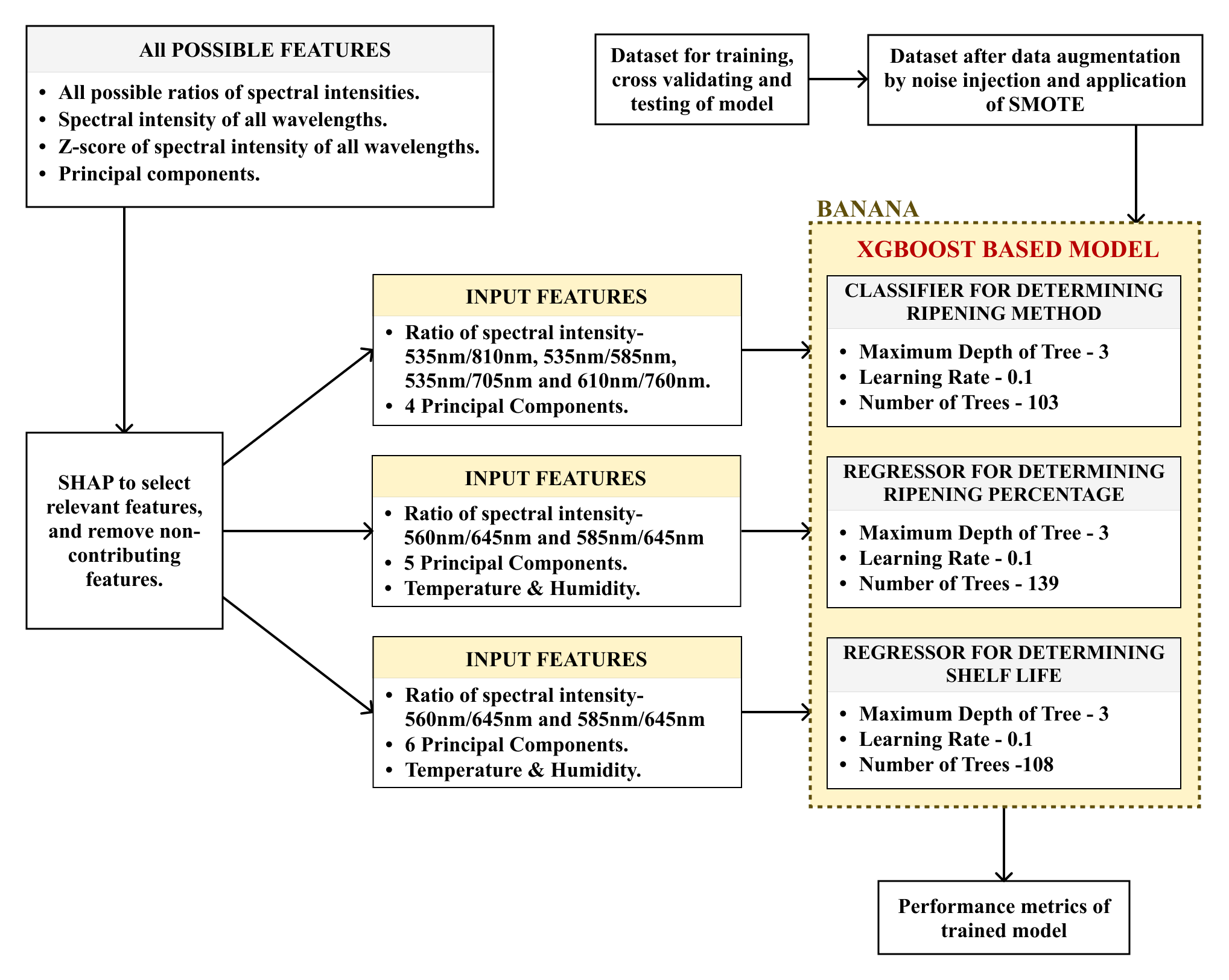} 
        \caption{}
        \label{fig:feature_set_banana}
    \end{subfigure}%
    \vspace{0.2cm} 
    \begin{subfigure}[b]{0.48\textwidth}
        \centering
        \includegraphics[width=1\textwidth, keepaspectratio]{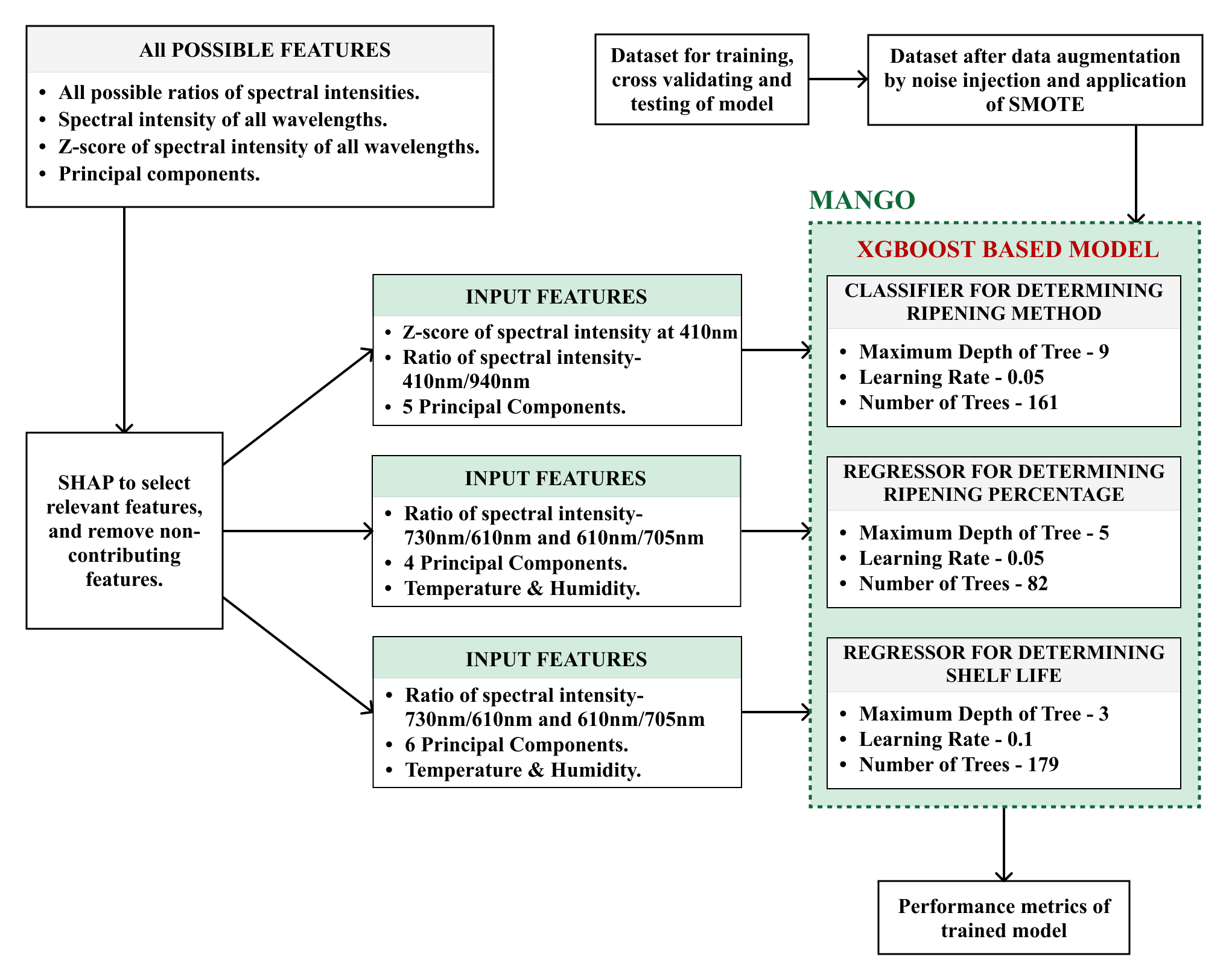}
        \caption{}
        \label{fig:feature_set_mango}
    \end{subfigure}%
    \caption{SHAP-based feature selection and XGBoost modeling pipelines for (a) banana and (b) mango samples. The schematic outlines the extraction of relevant input features using SHAP analysis on the training data after augmentation, and details the tuned hyperparameters (maximum depth, learning rate, and number of trees) used for classification and regression tasks.}
    \label{fig:model}
\end{figure}

Three independent models are trained, each for a specific predictive task. For the primary goal of identifying the ripening method, an XGBoost classifier is used, which classifies the fruit sample as either ‘Safely’ Ripened (consists of both natural and ethylene ripened classes) and ‘Carbide’ Ripened. The classifier pipeline includes a Synthetic Minority Over-sampling Technique (SMOTE) layer, for ensuring balance in support of both classes.
In addition to the classification model, two XGBoost regression models are trained to predict the remaining shelf life (in days) and the ripening progression (in percentage). The hyperparameters for each model are optimized independently using a comprehensive GridSearchCV strategy to ensure optimal predictive performance. Furthermore, the modeling pipelines are fruit-specific, meaning separate architectures are trained for mango and banana samples. Based on eigenvalue analysis and the obtained loading plots, the first five to seven principal components are used as input features; for both classification and regression tasks. These PCs combined with relevant wavelength ratios, temperature, humidity, and z-score of spectral intensity at 410 nm (for mango classifier only), are used to train the learning algorithms. The entire feature sets used are detailed in Figure~\ref{fig:model}.

\subsection{\textbf{Model Benchmarking and Performance Metrics}}
To validate the efficacy of the proposed XGBoost architecture, a comparative analysis is performed against three standard machine learning baselines, consisting of support vector machines (SVM), Decision Tree, and multi-layer perceptron (MLP) neural networks.
All these models are trained on the same spectral features, as mentioned in subsection 3.5. During 5-fold group cross-validation, the XGBoost classification model demonstrated strong stability across both fruit subsets, yielding a mean accuracy of $83.6\% \pm 4.0\%$ for mango and $77.5\% \pm 2.9\%$ for banana. When evaluated on the unaugmented hold-out test set, the XGBoost model achieved a classification accuracy of 95\% for mango samples and 81\% for banana samples, representing the highest overall accuracy among the evaluated classifiers.
The ‘Recall’ of carbide class is a crucial metric, as it is a measure of the model’s ability to detect calcium carbide-ripened samples correctly. SVM and Neural Network obtain a recall of 0.70 for mango, slightly better than the proposed XGBoost (0.67) and Decision Tree (0.60). SVM has a higher carbide recall of 0.80 for banana samples, while XGBoost gets 0.74, which is better than the decision tree (0.60) and neural network (0.55).
% --- TABLE 1: MANGO ---
\begin{table}[H]
    \centering
    \caption{Comparative analysis of evaluated models for classification of ripening method and shelf-life prediction in Mango (\textit{Mangifera indica}). Metrics include classification accuracy, carbide class recall, and regression coefficients ($R^2$, RMSE).}
    \label{tab:mango_fixed}
    \resizebox{\columnwidth}{!}{%
    \begin{tabular}{l c c c c}
        \toprule
         & \multicolumn{2}{c}{\textbf{Classification}} & \multicolumn{2}{c}{\textbf{Regression}} \\
        \cmidrule(lr){2-3} \cmidrule(l){4-5}
        \textbf{Model Architecture} & \textbf{Accuracy} & \textbf{Carbide Recall} & \textbf{Ripeness ($R^2$)} & \textbf{Shelf Life (RMSE)} \\
        \midrule
        SVM & 0.81 & 0.70 & 0.810 & 1.80 \\
        Decision Tree & 0.79 & 0.60 & 0.729 & 2.27 \\
        Neural Network (MLP) & 0.70 & 0.70 & 0.677 & 1.95 \\
        \midrule
        \textbf{XGBoost (5-Fold CV)} & \textbf{0.836 $\pm$ 0.040} & \textbf{--} & \textbf{0.659 $\pm$ 0.136} & \textbf{--} \\
        \textbf{XGBoost (Hold-out)} & \textbf{0.95} & \textbf{0.67} & \textbf{0.750} & \textbf{1.86} \\
        \bottomrule
    \end{tabular}%
    }
\end{table}
% --- TABLE 2: BANANA ---
\begin{table}[H] 
    \centering
    \caption{Comparative analysis of evaluated models for classification of ripening method and shelf-life prediction in Banana (\textit{Musa acuminata}). Metrics include classification accuracy, carbide class recall, and regression coefficients ($R^2$, RMSE).}
    \label{tab:banana_analysis}
    \resizebox{\columnwidth}{!}{%
    \begin{tabular}{l c c c c}
        \toprule
         & \multicolumn{2}{c}{\textbf{Classification}} & \multicolumn{2}{c}{\textbf{Regression}} \\
        \cmidrule(lr){2-3} \cmidrule(l){4-5}
        \textbf{Model Architecture} & \textbf{Accuracy} & \textbf{Carbide Recall} & \textbf{Ripeness ($R^2$)} & \textbf{Shelf Life (RMSE)} \\
        \midrule
        SVM & 0.79 & 0.80 & 0.850 & 1.36 \\
        Decision Tree & 0.73 & 0.60 & 0.779 & 1.62 \\
        Neural Network (MLP) & 0.55 & 0.55 & 0.847 & 1.39 \\
        \midrule
        \textbf{XGBoost (5-Fold CV)} & \textbf{0.775 $\pm$ 0.029} & \textbf{--} & \textbf{0.864 $\pm$ 0.055} & \textbf{--} \\
        \textbf{XGBoost (Hold-out)} & \textbf{0.81} & \textbf{0.74} & \textbf{0.871} & \textbf{1.30} \\
        \bottomrule
    \end{tabular}%
    }
\end{table}

\begin{figure}[H]
    \centering
    \begin{subfigure}[b]{0.48\textwidth}
        \centering
        \includegraphics[width=1\textwidth, keepaspectratio]{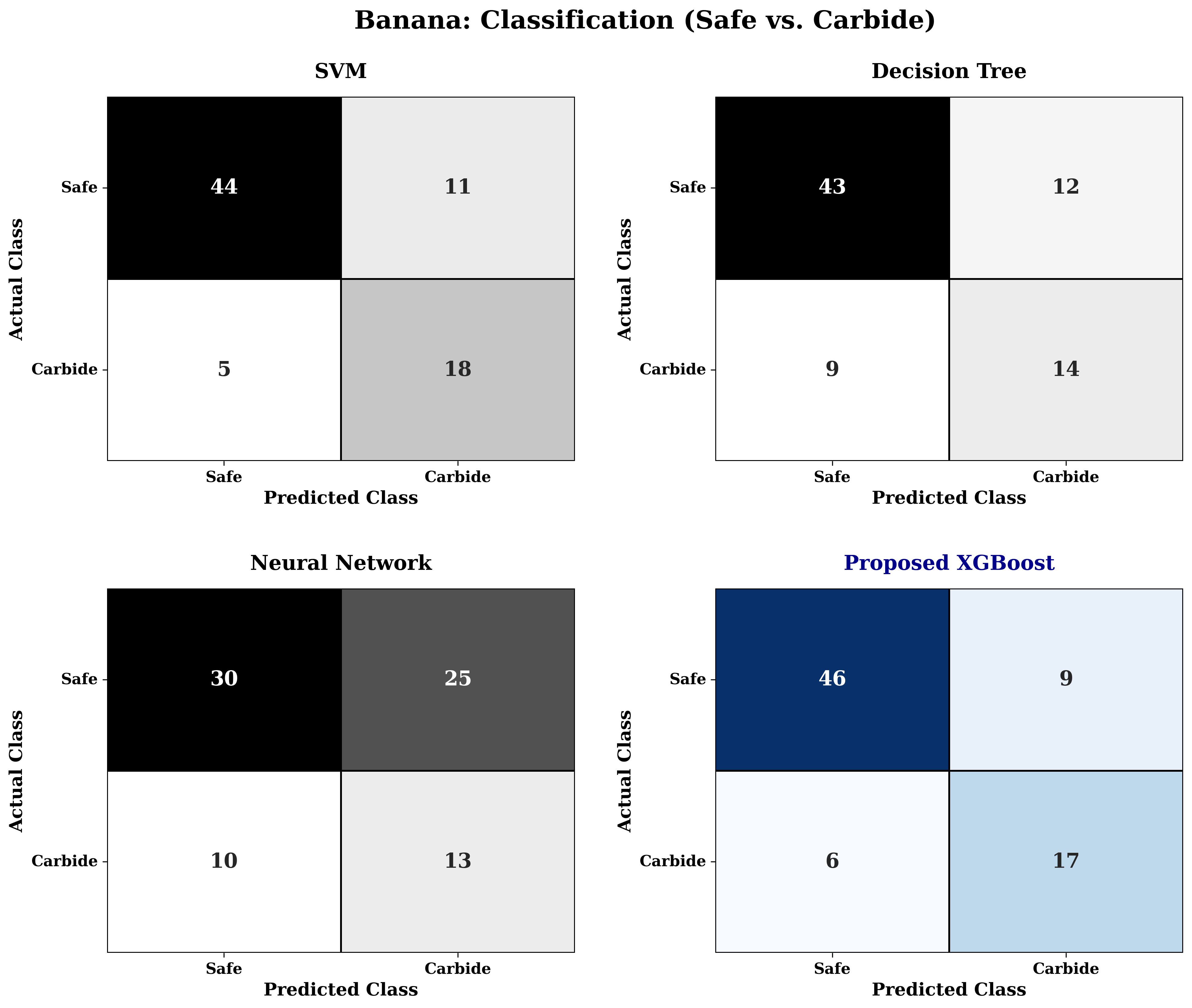} 
        \caption{}
        \label{fig:confusion_matrix_banana}
    \end{subfigure}%
    \vspace{0.2cm} 
    \begin{subfigure}[b]{0.48\textwidth}
        \centering
        \includegraphics[width=1\textwidth, keepaspectratio]{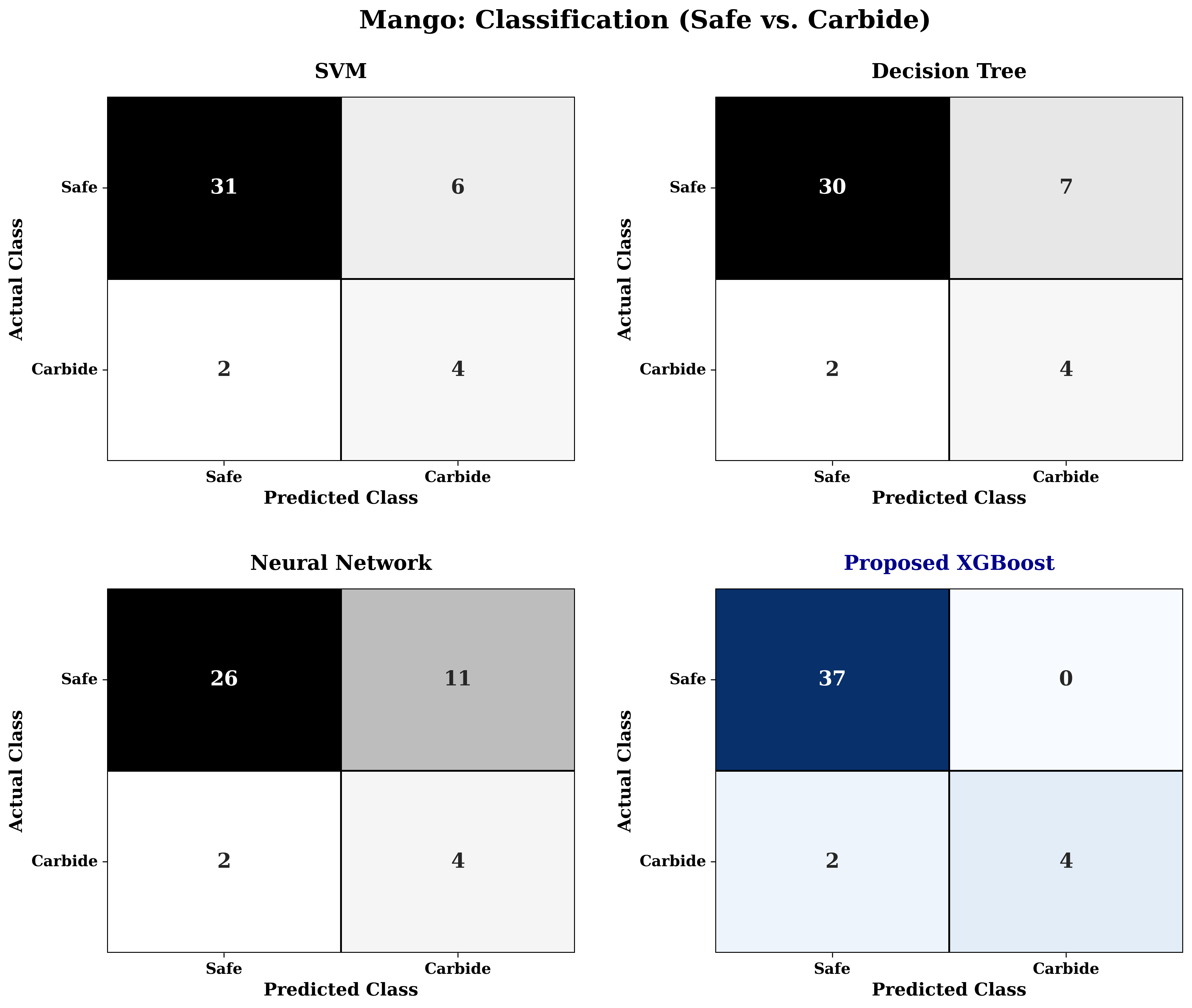}
        \caption{}
        \label{fig:confusion_matrix_mango}
    \end{subfigure}%
    \caption{Confusion matrices illustrating the classification performance of the four evaluated models (SVM, Decision Tree, Neural Network, and the Proposed XGBoost) for identifying calcium carbide-induced ripening in (a) Banana and (b) Mango samples. XGBoost achieves the highest overall classification accuracy, while SVM provides higher recall for the calcium carbide class.}
    \label{fig:classification_performance}
\end{figure}

\begin{figure}[H]
    \centering
    \begin{subfigure}[b]{0.48\textwidth}
        \centering
        \includegraphics[width=1\textwidth, keepaspectratio]{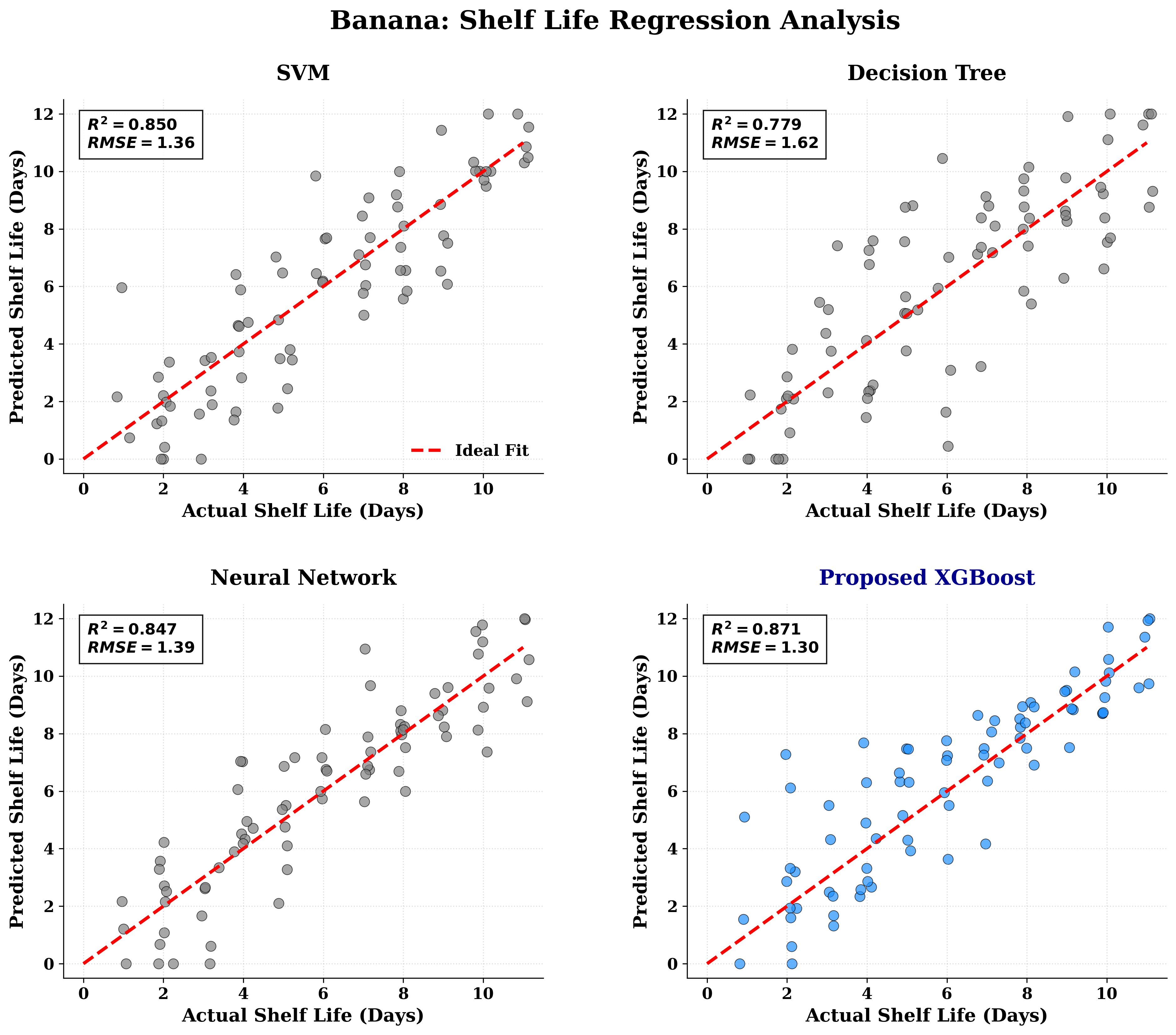} 
        \caption{}
        \label{fig:regression_banana}
    \end{subfigure}% 
    \vspace{0.2cm} 
    \begin{subfigure}[b]{0.48\textwidth}
        \centering
        \includegraphics[width=1\textwidth, keepaspectratio]{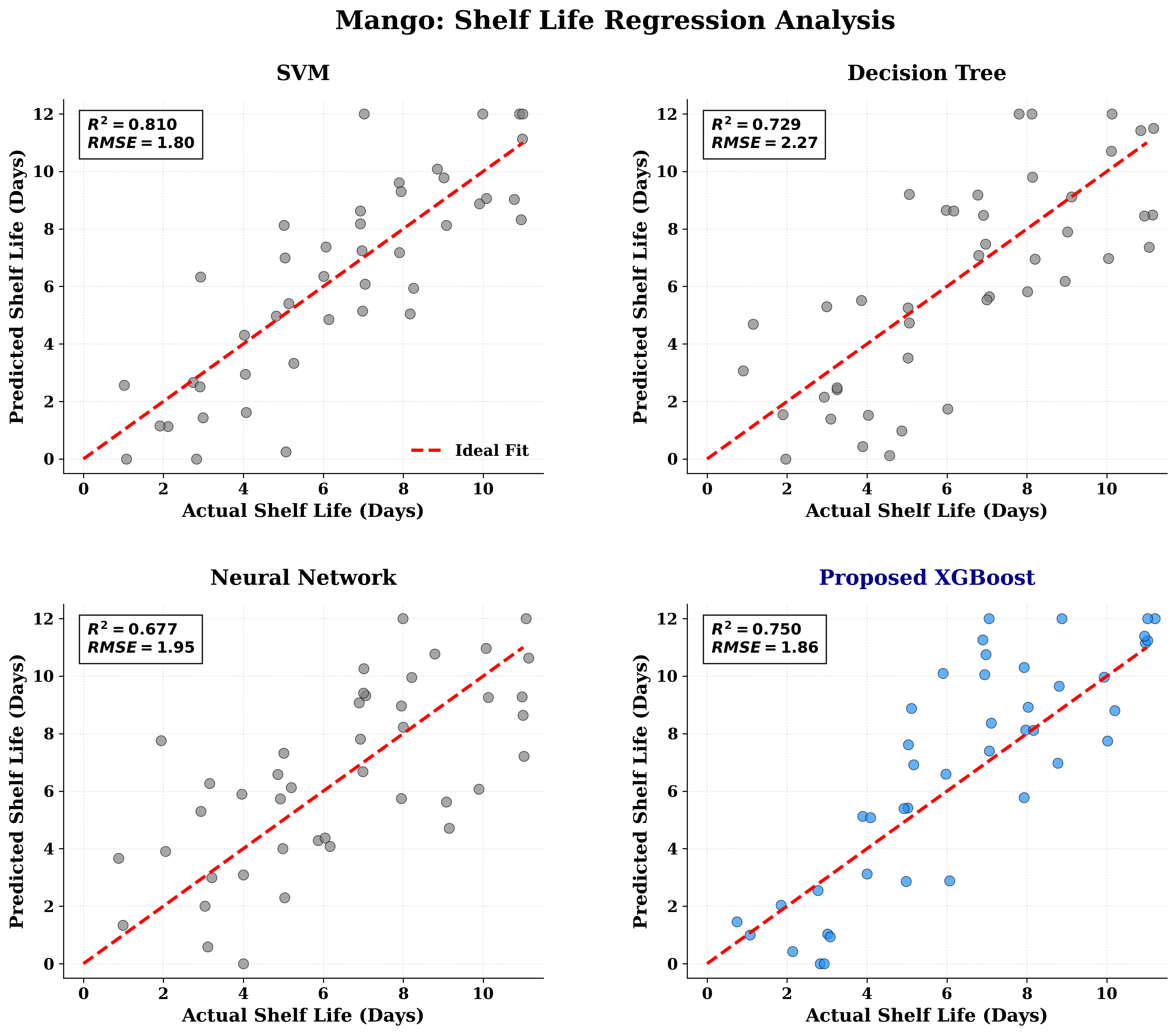}
        \caption{}
        \label{regression_mango}
    \end{subfigure}%
    \caption{A comparative analysis of regression performance of the four evaluated models (SVM, Decision Tree, Neural Network, and the Proposed XGBoost) in estimating the ripening progression in (a) Banana and (b) Mango samples. The XGBoost model demonstrates a tight convergence to the ideal fit line ($R^2 = 0.871$ for banana samples and $R^2 = 0.750$ for mango samples). This suggests that the gradient boosting framework captures a substantial portion of the non-linear spectral shifts associated with physiological changes in fruits over time.}
    \label{fig:regression_performance}
\end{figure}

While SVM has a slight advantage in recall, the proposed XGBoost architecture is preferred for classification as it shows significantly higher overall accuracy (95\% for mango and 81\% for banana), balancing the detection of safely and carbide ripened samples while minimizing the total number of misclassifications. (Table~\ref{tab:mango_fixed} and Table~\ref{tab:banana_analysis}). 
For regression models predicting ripeness percentage and remaining shelf-life, a nuanced performance divergence is observed between the fruits. In the case of mango samples, SVM actually outperforms the proposed XGBoost architecture, achieving a higher $R^2$ (0.810 vs. 0.750) and a lower shelf-life RMSE (1.80 vs. 1.86 days). Conversely, for banana samples, XGBoost reclaims its position as the most robust model, yielding an $R^2$ of 0.871 and an RMSE of 1.30 days, outperforming SVM (0.850 and 1.36 days, respectively). A visual description of the classification performance is provided in Figure~\ref{fig:classification_performance} in the form of confusion matrices. XGBoost achieved the highest overall classification accuracy among the evaluated classifiers, although SVM provided higher recall for the calcium carbide class. In Figure~\ref{fig:regression_performance}, the tight clustering of data points along the ideal fit line in the XGBoost plot demonstrates the model's robustness in handling the changes in spectral profile across the 11-day observation period. Despite SVM's highly competitive performance, particularly in mango regression, XGBoost is selected as the primary architecture for this study. Unlike SVMs and Neural Networks, which act as "black-box" models \citep{YANG202229}, the gradient boosting framework allows for explicit SHAP-based feature interpretability. This transparency is crucial for tracing non-linear spectral shifts back to specific physiological changes \citep{ijms27041805}.

\section{Conclusion and Future Scope}
The proposed study presents a comprehensive multispectral framework for non-invasive assessment of fruit ripening, with particular focus on distinguishing between safely ripened (including natural and ethylene ripened) and calcium carbide-ripened samples in mango (Mangifera indica) and banana (Musa acuminata). By analyzing spectral intensity data across 18 wavelengths (410–940 nm), this study demonstrates that distinct, method-specific spectral signatures emerge at various maturation stages. Temporal progression of intensity values reveals clear differences between naturally ripened and calcium carbide-ripened samples, where calcium carbide-treated samples exhibit stronger spectral changes in the visible region, which are consistent with faster chlorophyll breakdown and carotenoid development. However, this is not accompanied by a proportionate change in the measured NIR region response, suggesting that the internal maturation of the fruit does not align with the rapid surface color transformation. Dimensionality reduction of the spectral intensity values at all 18 wavelengths using PCA reveals a low dimensional feature space where first 5-7 principal components capture >90\% of the spectral variance. These principal components, combined with targeted spectral intensity ratios and environmental parameters (temperature and relative humidity), form a robust feature set. This set is subsequently utilized to train XGBoost-based learning algorithms for classification of ripening method, predicting the remaining shelf-life in days and estimating the ripening progression. The proposed study achieves a classification accuracy of 95\% for mango samples with a recall of 0.67 for calcium carbide-ripened samples. Similarly, banana samples are classified with an accuracy of 81\% and calcium carbide class recall of 0.74. This study provides a strong initial validation of the proposed
multispectral framework for non-invasive assessment of fruit ripening under controlled experimental conditions. While the results obtained are promising, broader validation on additional cultivars, independent fruit batches, and varied environmental conditions would strengthen the generalizability of the approach.
Future work may focus on extending the framework to other climacteric fruits, integrating larger multi-season datasets, and validating the findings with biochemical and physiological ground-truth measurements. Such extensions will support a more robust deployment of the system in real-world post-harvest monitoring applications.
\section*{Statements and Declarations}
\subsection*{\textbf{Data Availability}}
The dataset collected and analyzed during this study is available from the corresponding author on reasonable request.

\subsection*{\textbf{Author Contributions}}
\textbf{Gurbhit Chaurakoti:} Conceptualization, Methodology, Investigation, Data curation, Formal analysis, Visualization, Software, Writing – original draft, Writing – review \& editing. \textbf{Harshit Kumar:} Conceptualization, Methodology, Resources, Investigation, Visualization, Software, Validation, Writing – original draft, Writing – review \& editing. \textbf{Hani Kumar:} Data curation, Investigation, Visualization, Writing – review \& editing. \textbf{Anurag Singh:} Supervision, Project administration, Methodology, Writing – review \& editing. \textbf{Ram Asrey:} Conceptualization, Validation, Writing – review \& editing.

\subsection*{\textbf{Declaration of competing interest}}
The authors have no relevant financial or non-financial interests to disclose.

\subsection*{\textbf{Funding}}
This work is supported by IIT Ropar Technology and Innovation Foundation, one of the
Technology Innovation Hubs under DST NM-ICPS (National Mission on Interdisciplinary Cyber-Physical Systems), Government of India (Ref.No. AWaDH/2025/SAMRIDHI4.0/Ideathon/TI06).
\nocite{*}
\bibliographystyle{elsarticle-harv} 
\bibliography{myreferences}
%%%%%%%%%%%%%%%%%%%%%%%%%%%%%%%%%%%%%%%%%%%%%%%%%%%%%%%%%%%%%%%

\end{document}